\documentclass[11pt]{article}
\usepackage{xcolor}

\usepackage[final]{acl}
\usepackage{booktabs}
\usepackage{amsmath}
\usepackage{multirow}
\usepackage{pifont} 
\usepackage[skins,breakable]{tcolorbox}
\usepackage{xcolor}
\usepackage{times}
\usepackage{latexsym}
\usepackage{array}
\usepackage[table]{xcolor} 
\newcommand{\revise}[1]{\textcolor{black}{#1}}
\usepackage{multirow}
\usepackage{fontawesome5}
\usepackage[T1]{fontenc}
\usepackage{booktabs}
\usepackage{tabularx}
\usepackage[table]{xcolor}
\definecolor{CompressColor}{HTML}{E8F4FF} 
\definecolor{GroundColor}{HTML}{FFF2D9} 
\usepackage[utf8]{inputenc}

\usepackage{microtype}

\usepackage{inconsolata}

\usepackage{graphicx}

\title{DVBench: Benchmarking MLLMs for Understanding Dynamic Charts and Narratives in Data Videos}

\author{
Bomiao Wang$^{1}$\thanks{Equal contribution; co-first authors.},
Zekai Shao$^{1}$\footnotemark[1],
Jiexiang Lan$^{1}$,
Xiaoliang Fu$^{1}$,
Xingchen Zeng$^{2}$,
Siming Chen$^{1}$\thanks{Corresponding author.}
\\
$^{1}$Fudan University,
$^{2}$The Hong Kong University of Science and Technology (Guangzhou)
\\
\texttt{rimeiw66@gmail.com},
\texttt{zkshao23@m.fudan.edu.cn},
\texttt{simingchen@fudan.edu.cn}
}

\begin{document}
\maketitle
\begin{abstract}
While MLLMs have made significant strides in chart comprehension and video understanding, current evaluations largely isolate these capabilities, leaving a critical gap in understanding temporally evolving structured visual information. To address this gap, we introduce \textbf{DVBench}, a benchmark for evaluating MLLMs on data videos, a storytelling medium that integrates dynamic charts with structured narratives. We decompose data video understanding into five dimensions. DVBench comprises 300 real-world data videos and 1,000 human-verified QA pairs curated through a rigorous semi-automated pipeline. Extensive evaluations of nine MLLMs show that Gemini-3.1-Pro achieves the best overall performance, while Kimi-k2.5 is the strongest open-source model. We further identify two notable phenomena: open-source model performance does not scale strictly with parameter size, and narrative proficiency does not guarantee visual capability. \revise{Fine-grained analyses and ablation studies further reveal dimension-specific weaknesses and the effects of frame configurations and subtitle inputs, informing future MLLM development.
DVBench is publicly available at \url{https://bomiaowang.github.io/DVBench/}.}
\end{abstract}

\begin{figure*}[!ht]
    \centering
    \includegraphics[width=\textwidth]{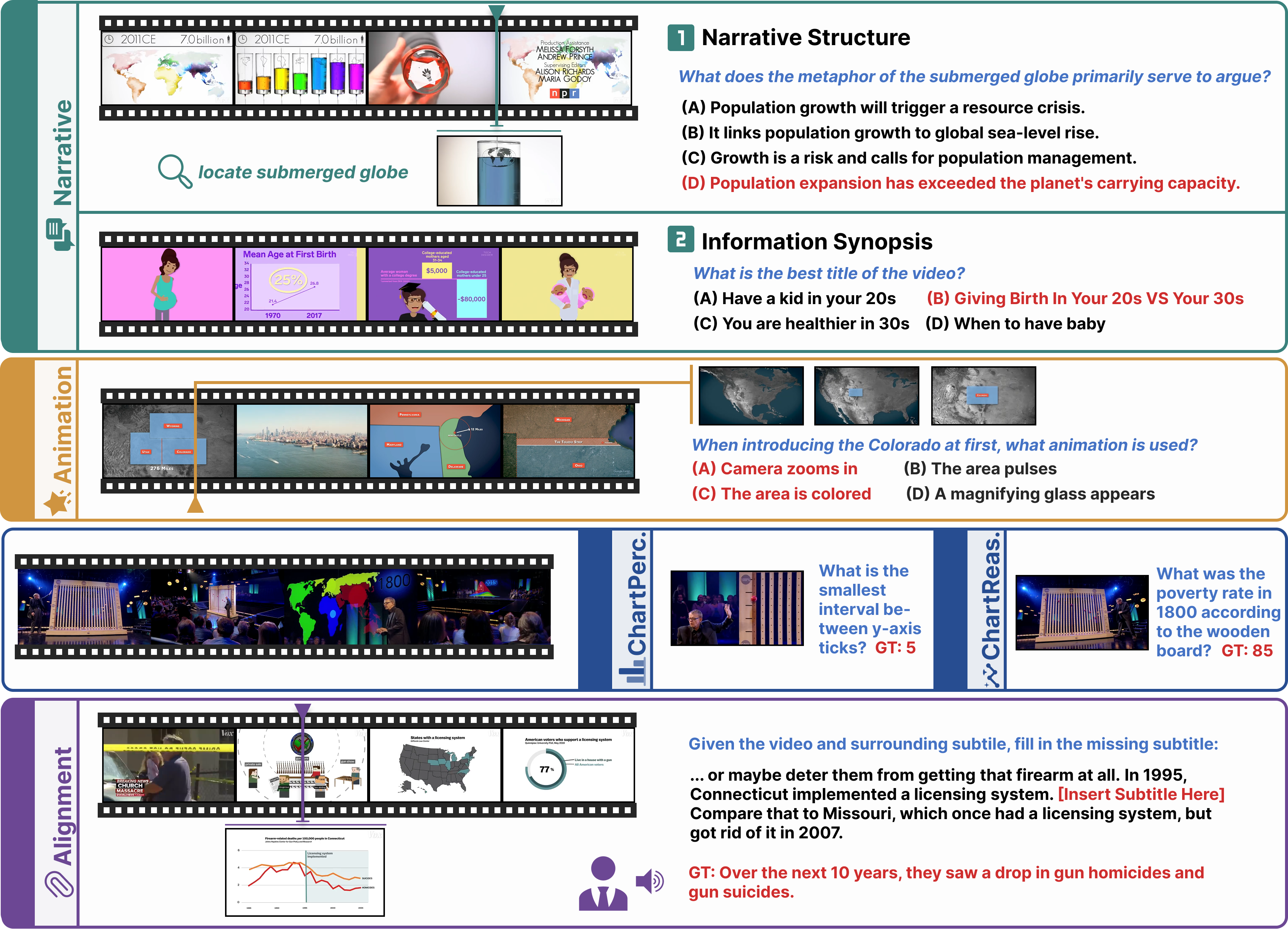}
    \caption{\textbf{Overview of the five evaluation dimensions in DVBench}. Our taxonomy systematically assesses models across: (1) \textbf{Narrative}, evaluating high-level comprehension of video themes; (2) \textbf{Animation}, focusing on the perception of dynamic transitions; (3) \textbf{Chart Perception} (denoted as ChartPerc.) and (4) \textbf{Chart Reasoning} (denoted as ChartReas.), respectively test fundamental visual identification and data insights extraction from dynamic charts in videos; and (5) \textbf{Alignment}, requiring models to infer masked subtitles. Correct options and ground truth (GT) answers are colored in red.}
    \label{fig:task_def}
\end{figure*}

\section{Introduction}
The continuous evolution of MLLMs has significantly advanced the frontier of visual understanding, enabling breakthroughs in cross-modal comprehension and complex reasoning~\cite{zhang2024llava, google2025gemini3blog, anthropic2026claude46}. To comprehensively assess the capabilities and progress of MLLMs, existing benchmarks evaluate models from multiple aspects of visual understanding~\cite{li2024survey}, among which chart understanding evaluates a model's ability to decode structured, dense visual information and perform precise analytical reasoning~\cite{masry2022chartqa, wang2024charxiv}, while video understanding assesses its capacity to track dynamic entities and comprehend sequential events~\cite{li2024mvbench, fu2025videomme}.

However, these two foundational lines of evaluation have remained largely isolated, leading to a critical blind spot in current benchmarks: the perception and reasoning of dynamic, structural visual information over time. On one hand, chart understanding benchmarks are restricted to static visualizations~\cite{wang2024charxiv, xia2025chartx, masry2022chartqa}. On the other hand, video understanding benchmarks focus almost exclusively on physical entities and actions within natural scenes. Even when complex narratives are introduced, they are typically confined to cinematic or real-world events~\cite{wang2025lvbench, zhang2025seriesbench}, leaving a significant gap in the comprehension of data-driven storytelling.

This distinct fragmentation underscores the necessity of evaluating models on \textit{Data Videos} (see Fig.~\ref{fig:task_def} for an illustrative example), a storytelling medium that bridges this gap by integrating dynamic visualizations, narrative structures, and cinematic transitions~\cite{heer2007animated}. \revise{
Understanding data videos poses a comprehensive challenge for MLLMs, requiring the ability to (1) read and interpret dynamic charts, including their visual encodings, data insights, and animation semantics; (2) understand how these insights are organized and developed into a coherent narrative; and (3) integrate the correspondence between visual content and textual narration.
}

To this end, we introduce \textbf{DVBench}, the first comprehensive benchmark dedicated to evaluating the comprehension of data videos across five dimensions: \textbf{Narrative}, \textbf{Animation}, \textbf{Chart Perception}, \textbf{Chart Reasoning}, and \textbf{Alignment}. DVBench comprises 300 high-quality data videos, accompanied by 1,000 QA pairs constructed through a rigorous semi-automatic pipeline and human verification.

With this benchmark, we conduct extensive evaluations of nine prominent open-source and proprietary MLLMs. The experimental results show Gemini-3.1-Pro achieves the best performance, and Kimi-k2.5 emerges as the leading open-source model. Furthermore, we observe two notable phenomena: first, the capacity to understand data videos does not scale strictly with the model size; second, proficiency in the narrative understanding does not correlate with performance in the fine-grained visual perception. 

To further characterize model capabilities, we conduct fine-grained analyses using question-level annotations derived from the design space of data videos. The results reveal model weaknesses in perceiving non-linear animation pacing and counting in dynamic charts, while robustness to increasing video length varies across models. \revise{Ablations further show that denser frame sampling does not necessarily help models, and incorporating subtitles leads to performance gains.}

Our contributions are as follows: (1) we introduce a new task, data video understanding, with five dimensions to systematically evaluate MLLMs; (2) we curate a dataset of 300 data videos and 1,000 QA pairs via a rigorous semi-automatic pipeline and manual review; and (3) we conduct extensive evaluations with fine-grained analyses and ablations, yielding insights into current MLLM capabilities and limitations.

\section{Related Works}
We compare DVBench and other multimodal understanding benchmarks in Table~\ref{tab:related_work_comparison}. The following subsections discuss the current benchmarks of chart and video understanding and data video study, highlighting the necessity of our DVBench.
\begin{table}[t]
\centering
\setlength{\tabcolsep}{3pt} 
\resizebox{\columnwidth}{!}{
\begin{tabular}{lcccc}
\toprule
\textbf{Benchmark} & \textbf{Dynamic Chart} & \textbf{Narratives} & \textbf{Annotation} & \textbf{Multi-dim. Eval} \\
\midrule
ChartQA~\cite{masry2022chartqa} & \textcolor{red}{\ding{55}} & \textcolor{red}{\ding{55}} & {\small\faUser} \,\&\, {\small\faRobot} & \textcolor{red}{\ding{55}} \\
CharXiv~\cite{wang2024charxiv} & \textcolor{red}{\ding{55}} & \textcolor{red}{\ding{55}} & {\small\faUser} \,\&\, {\small\faRobot} & \textcolor{red}{\ding{55}} \\
ChartX~\cite{xia2025chartx} & \textcolor{red}{\ding{55}} & \textcolor{red}{\ding{55}} & {\small\faUser} \,\&\, {\small\faRobot} & \textcolor{green}{\ding{51}} \\
VisText~\cite{tang2023vistext} & \textcolor{red}{\ding{55}} & \textcolor{green}{\ding{51}} & {\small\faUser} \,\&\, {\small\faRobot} & \textcolor{red}{\ding{55}} \\
ChartInsighter~\cite{wang2025chartinsighter} & \textcolor{red}{\ding{55}} & \textcolor{green}{\ding{51}} & {\small\faUser} & \textcolor{green}{\ding{51}} \\ 
ChartCap~\cite{lim2025chartcap} & \textcolor{red}{\ding{55}} & \textcolor{green}{\ding{51}} & {\small\faUser} \,\&\, {\small\faRobot} & \textcolor{red}{\ding{55}} \\
\midrule
MVBench~\cite{li2024mvbench} & \textcolor{red}{\ding{55}} & \textcolor{red}{\ding{55}} & {\small\faRobot} & \textcolor{green}{\ding{51}} \\
MotionBench~\cite{hong2025motionbench} & \textcolor{red}{\ding{55}} & \textcolor{red}{\ding{55}} & {\small\faUser} & \textcolor{red}{\ding{55}} \\
Video-MMMU~\cite{hu2025videommmu} & \textcolor{red}{\ding{55}} & \textcolor{red}{\ding{55}} & {\small\faUser} & \textcolor{green}{\ding{51}} \\
LVBench~\cite{wang2025lvbench} & \textcolor{red}{\ding{55}} & \textcolor{green}{\ding{51}} & {\small\faUser} \,\&\, {\small\faRobot} & \textcolor{green}{\ding{51}} \\
VRBench~\cite{yu2025vrbench} & \textcolor{red}{\ding{55}} & \textcolor{green}{\ding{51}} & {\small\faUser} \,\&\, {\small\faRobot} & \textcolor{green}{\ding{51}} \\
SeriesBench~\cite{zhang2025seriesbench} & \textcolor{red}{\ding{55}} & \textcolor{green}{\ding{51}} & {\small\faUser} \,\&\, {\small\faRobot} & \textcolor{green}{\ding{51}} \\
\midrule
\textbf{DVBench (Ours)} & \textcolor{green}{\ding{51}} & \textcolor{green}{\ding{51}} & {\small\faUser} \,\&\, {\small\faRobot} & \textcolor{green}{\ding{51}} \\
\bottomrule
\end{tabular}
}
\caption{\textbf{Comparison of DVBench with existing multimodal understanding benchmarks.} {\small\faUser} and {\small\faRobot} respectively denote manual and automatic dataset curation. Multi-dim. Eval indicates whether the benchmark assesses multi-dimensional capabilities. 
}
\label{tab:related_work_comparison}
\end{table}

\subsection{Chart Understanding Benchmarks}
Current benchmarks in chart understanding mainly include chart QA and chart captioning. Early chart QA benchmarks focused on structural extraction and reasoning over scientific data within open-vocabulary scenarios~\cite{kafle2018dvqa, methani2020plotqa}. Subsequent works introduced more sophisticated reasoning logic and expert-validated questions covering diverse real-world topics~\cite{masry2022chartqa, wang2024charxiv, wu2024chartinsights, xia2025chartx}. Evaluation frameworks for chart captioning are dedicated to faithfulness, granularity, and hallucination-free generation~\cite{tang2023vistext, wang2025chartinsighter, lim2025chartcap, wang2026chartfi}. However, these benchmarks all remain confined to static visualizations. In contrast, our work investigates dynamic charts within videos, which pose a significantly higher cognitive challenge.

\subsection{Video Understanding Benchmarks}
The evaluation of video understanding capability is currently shifting from short clips toward long-form content and complex narratives. Early benchmarks progressed from static spatial perception to the dynamic perception of actions and object state changes~\cite{li2024mvbench, hong2025motionbench, ha2026narrativetrack}. Subsequent research introduced reasoning tasks designed to evaluate causal relationships and temporal logic within long videos~\cite{wang2025lvbench, yu2025vrbench}. Regarding video caption, recent frameworks focus on fine-grained evaluation by quantifying correctness, comprehensiveness, and instruction-following capabilities~\cite{li2025capfollowins, liu2025goodcomprehensivecap}. Furthermore, various domain-specific video benchmarks have emerged to evaluate specialized narratives, ranging from cinematic plot progression~\cite{shah2025cin, ataallah2025infinibench} to multi-disciplinary knowledge and scientific processes~\cite{hu2025videommmu, deng2025scivideobench}. Our DVBench establishes a framework to evaluating the understanding of data-driven narratives.

\subsection{Data Video}
Data videos have emerged as a prominent storytelling medium across diverse domains, demonstrating an exceptional ability to engage audiences and convey rich information through the integration of dynamic visualizations and structured narratives~\cite{segel2010narrative, amini2015understanding, narrativeplayer, shen2025reflecting}. Extensive research has investigated the foundational elements, characterizing the design space of visual-animation interplay and scene-semantic alignment~\cite{shi2021communicating, cheng2022investigating, gao2025sceneloom}. Furthermore, researchers have investigated narrative strategies to better organize and convey information in data videos~\cite{yang2021freytagPyramid, xu2022wow, xu2023end, wei2024telling}. Building on the research, we propose a systematic evaluation framework to assess MLLMs across diverse dimensions.

\section{DVBench}
\subsection{Task Dimensions}
\label{sec:task_dim}
Inspired by research in data videos, we deconstruct the understanding of data videos into five dimensions: Narrative, Animation, Chart Perception, Chart Reasoning, and Alignment in Fig~\ref{fig:task_def}.

For \textbf{Animation}, \textbf{Chart Reasoning}, and \textbf{Alignment}, we further divide each dimension into fine-grained categories derived from prior data video research~\cite{wang2019datashot, shi2021communicating, cheng2022investigating}. These categories characterize different capabilities and support the analyses in Sec.~\ref{sec:analysis}. Definitions are introduced below.

\textbf{Narrative} assesses the ability to comprehend the high-level storytelling logic and communicative intent. It is further divided into two tasks: (1) \textit{Narrative Structure} focuses on \textit{how} the video narrates, which involves three question types: mapping visuals to narrative intent (e.g., What does the metaphor of the submerged globe primarily serve to argue?), mapping intent back to specific visuals (e.g., What visualization is presented to illustrate the contrast between public perception and reality?)~\cite{yang2021freytagPyramid}, and identifying the logical relationships between video clips (e.g., Why does the video introduce the income distribution chart by a world map?). (2) \textit{Information Synopsis} focuses on \textit{what} the video narrates, which involves tasks such as selecting the best title and identifying the video's attitude towards the topic.

\textbf{Animation} focuses on the perception of dynamic animation techniques and visual transitions used in charts. We characterize this dimension using four editorial layers~\cite{shi2021communicating}: (1) \textit{visualization elements} (e.g., transforming visual marks, axes, or legends), (2) \textit{added elements} (e.g., adding annotations and embellishments), (3) \textit{camera} (e.g., zooming or panning to alter viewing perspectives), and (4) \textit{timeline} (e.g., controlling the pacing and speed of animation).

\textbf{Chart Perception} evaluates the recognition and direct reading of visual encodings and explicitly presented graphical information, including chart titles, axis labels, legend entries, tick marks, the number of ticks, and tick interval sizes. \revise{The target information is explicitly presented in the chart and can be obtained through direct visual reading, without requiring the derivation of higher-level data insights.}

\revise{\textbf{Chart Reasoning} evaluates the ability to derive higher-level data insights through reasoning over the information presented in the chart. In contrast to Chart Perception, the target information is not explicitly stated as a graphical element and must be inferred by analyzing one or more visual elements, potentially across multiple frames.} We systematically categorize these insights into 10 fine-grained types according to~\cite{wang2019datashot}, ranging from localized queries (i.e., retrieving \textit{values}, identifying \textit{extremes}, and conditional \textit{categorization}) and comparative statistics (i.e., measuring \textit{proportions}, comparing \textit{differences}, \textit{ranking}, and \textit{aggregation}) to broader structural analyses (i.e., tracking temporal \textit{trends}, analyzing \textit{distributions}, and finding \textit{associations}).

\textbf{Alignment} evaluates the correspondence between dynamic charts and their narration through a subtitle cloze task, where models infer missing narrations from visual cues and surrounding subtitle context. We characterize this dimension using two semantic categories based on the role of the masked subtitle: \textit{Data Insight} and \textit{Data Context}. \textit{Data Insight} refers to analytical facts, such as temporal trends or exact values, that are grounded in explicit visual evidence, whereas \textit{Data Context} refers to background information that describes the broader narrative setting~\cite{cheng2022investigating}.

\subsection{Benchmark Construction}
\begin{figure*}[!ht]
    \centering
    \includegraphics[width=\textwidth]{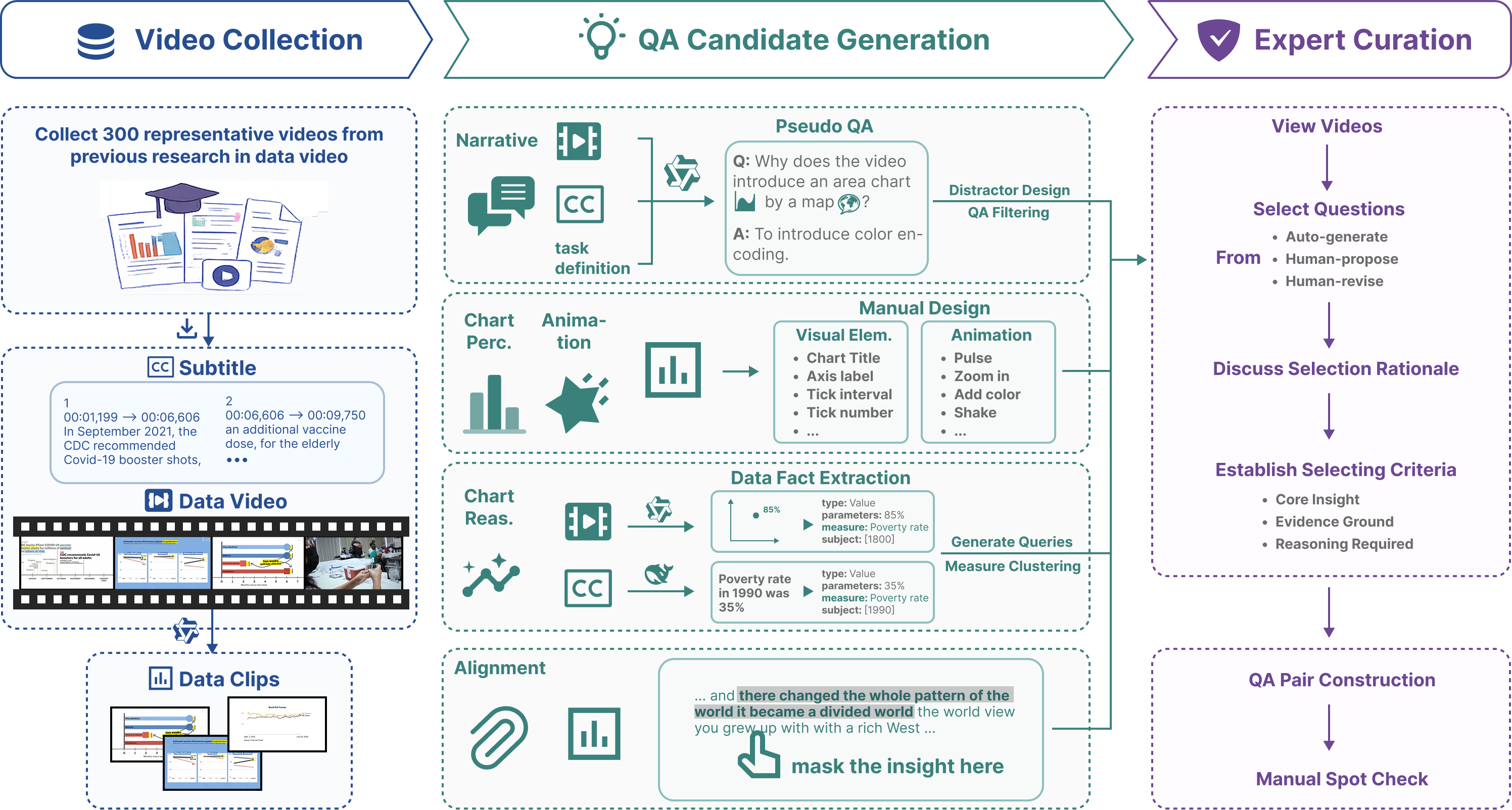}
    \caption{\textbf{Benchmark construction pipeline.} The process is divided into three main stages: (1) video collection, (2) QA candidate generation, and (3) expert curation. We collect videos from previous research and download the videos and subtitles. Then we utilize MLLM to identify data clips. During QA candidate generation, we employ dimension-specific strategies: Narrative questions are generated via MLLMs and undergo rigorous distractor design and filtering; Chart Perception and Animation questions are manually curated to target fundamental visual elements and dynamic transitions; Chart Reasoning utilizes an automated pipeline to extract, formalize, and cluster data insights for complex cross-frame queries; and Alignment questions are formulated as subtitle cloze tasks. Finally, annotators establish selection criteria and conduct benchmark curation and verification.}
    \label{fig:benchmark_construction}
\end{figure*}

 Our benchmark construction pipeline consists of three stages, as illustrated in Fig.~\ref{fig:benchmark_construction}.

\paragraph{Video Collection.} 
To ensure the quality and representativeness of our dataset, we collect 300 data videos from prior research dedicated to data videos~\cite{yang2021freytagPyramid, shi2021communicating, cheng2022investigating, xu2022wow, xu2023end, gao2025sceneloom, gunturu2025mapstory}. These videos cover a broad range of topics and presentation styles, while consistently featuring dynamic charts and structured narratives.

\paragraph{QA Pair Candidate Generation.}
We employ dimension-specific strategies to generate QA pair candidates.

For \textbf{Narrative} dimension, we input the video and subtitle into Qwen3-VL-Plus~\cite{bai2025qwen3vltechnicalreport} to generate pseudo questions for the two narrative tasks following the taxonomy defined in Sec.~\ref{sec:task_dim}. We further refine the distractors by, for example, mapping visual cues to irrelevant narrative strategies. Finally, we filter out trivial questions that Kimi-k2.5, Gemini-3.1-Pro, and Qwen3.5-397B-A17B can all answer correctly.

For the remaining four dimensions, we first employ Qwen3-VL-Plus to locate data clips, i.e., video segments in which charts convey data context or insights~\cite{amini2016authoring, cheng2022investigating}. Based on these identified clips, we construct QA candidates for dimensions below.

For \textbf{Animation} and \textbf{Chart Perception} dimensions, we manually design questions targeting dynamic transitions and fundamental visual elements within the data clips, respectively. For \textbf{Chart Reasoning} dimension, to ensure a comprehensive extraction of the underlying data insights, we simultaneously extract visual data insights from the data clips using Qwen3-VL-Plus and textual insights from the subtitles using DeepSeek-v3.2~\cite{deepseek2025v32}. After combining and formalizing these multi-modal insights into a structured schema (i.e., data fact)~\cite{wang2019datashot}, the automated pipeline generates two distinct types of questions. The first type directly queries the extracted data insights (e.g., identifying specific values, trends, or differences). To generate the second type, we further cluster the insights by their \textit{subjects} and \textit{measures}, utilizing these clusters to construct complex temporal tasks, such as tracking the change of a specific measure across different timestamps. For \textbf{Alignment} dimension, we use the identified data clips and mask narrations corresponding to either data insights or data contexts to construct subtitle cloze tasks. Detailed prompt templates used for QA candidate generation are provided in Appendix~\ref{sec:appendix_dataset_prompt}.

\revise{\paragraph{Expert Curation.}
To ensure annotation quality, two annotators with visualization expertise first jointly viewed 10 videos to establish a consistent annotation protocol. They independently selected questions from an initial pool of 88 candidates and then discussed the rationale behind their selections, from which they derived a shared question selection criteria. Following these criteria, one annotator constructed the full QA set, while the other independently inspected 10\% of the QA pairs, achieving 100\% agreement. Details are in Appendix~\ref{sec:appendix_annotation_interface}.}
\begin{figure*}[ht] 
    \centering
    \includegraphics[width=0.9\textwidth]{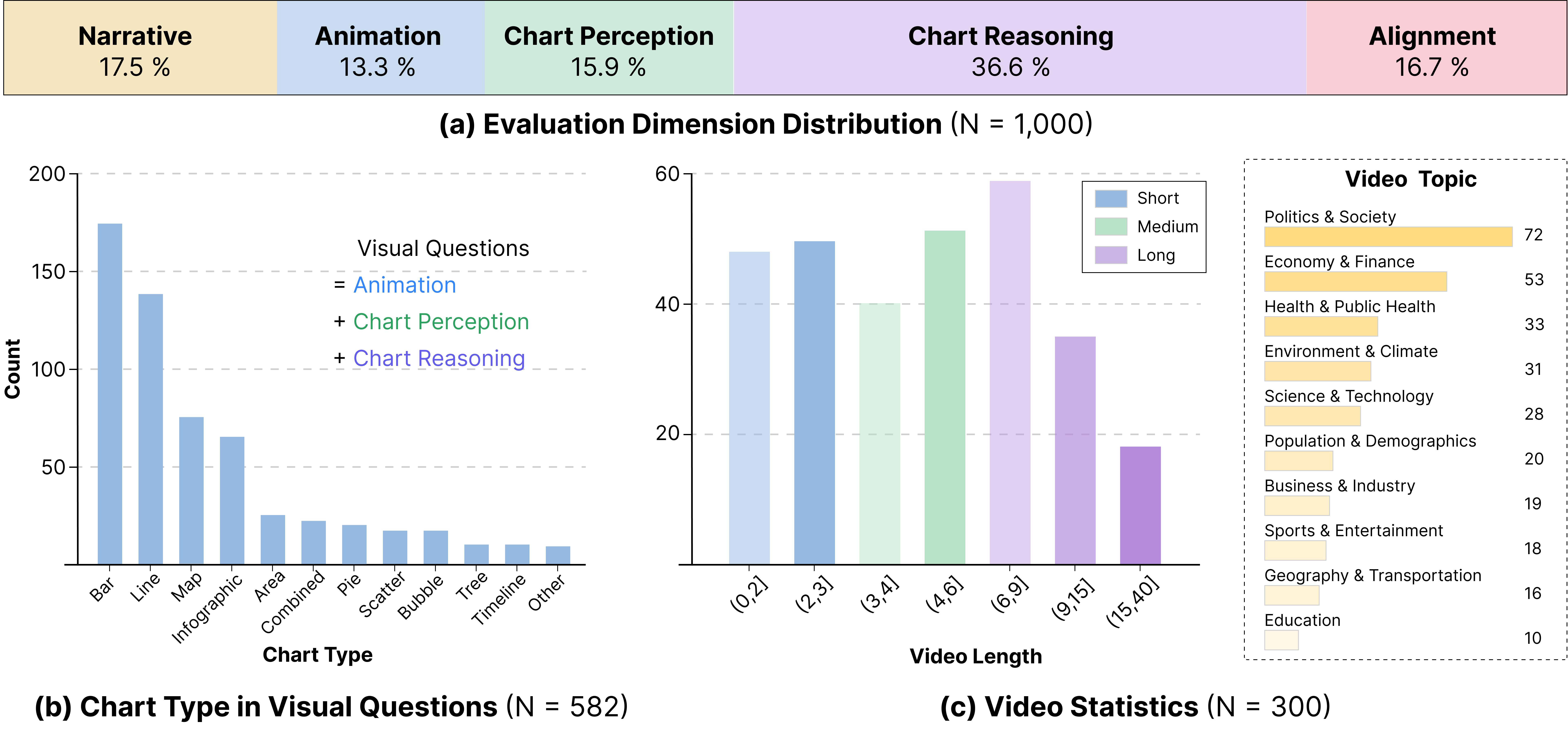}
    \caption{\revise{\textbf{Statistics of DVBench.} \textbf{(a)} Distribution of questions across the five evaluation dimensions. \textbf{(b)} Chart-type distribution of visual questions and its dimension-specific breakdown. Visual questions include only Animation, Chart Perception, and Chart Reasoning. \textbf{(c)} Video length and topic distribution of the collected data videos.}}
    \label{fig:dataset_statistics}
\end{figure*}

\subsection{Dataset Statistics}
\paragraph{QA Pairs.}
Fig.~\ref{fig:dataset_statistics}(a) details the distribution of the 1,000 QA pairs across our five evaluation dimensions. We allocate the largest proportion to Chart Reasoning for two reasons. First, viewers predominantly focus on the underlying data insights rather than superficial graphical elements. Second, prior study~\cite{wang2024charxiv} indicates that while models excel at basic perception, they still exhibit vulnerabilities in reasoning.

Figure~\ref{fig:dataset_statistics}(b) reports the chart type distribution in visual questions, which include Animation, Chart Perception, and Chart Reasoning dimensions only. Bar and line charts are the most prevalent chart types. The remaining questions cover a diverse long tail of diverse chart types, demonstrating the broad visual coverage of DVBench. Statistics on QA lengths are provided in Appendix~\ref{sec:appendix_dataset_statistics}.

\paragraph{Videos.}
Figure~\ref{fig:dataset_statistics}(c) presents the video length and topic distribution. Video lengths range from under 30 seconds to approximately 37 minutes. We group them into three duration categories: Short ($\leq 3$ mins), Medium ($3$--$6$ mins), and Long ($>6$ mins). The three groups are relatively balanced. \revise{The collected videos cover ten real-world topics, with Politics \& Society and Economy \& Finance being the two largest categories.}

\section{Experiment}
\begin{table*}[!ht]
\centering
\scriptsize
\setlength{\tabcolsep}{3pt}
\renewcommand{\arraystretch}{1.15}

\resizebox{\textwidth}{!}{
\begin{tabular}{l ccccc cccccc}

\toprule

\multirow{2}{*}{\textbf{Model}} &
\multicolumn{5}{c}{\textbf{Closed-ended Dimensions}} &
\multicolumn{6}{c}{\textbf{Align.}} \\

\cmidrule(lr){2-6}
\cmidrule(lr){7-12}

&
\textbf{Avg.} &
\textbf{Narr.} &
\textbf{Anim.} &
\textbf{Perc.} &
\textbf{Reas.} &
\textbf{BL-2} &
\textbf{MET} &
$\mathbf{F1_{\mathrm{BERT}}}$ &
\textbf{EMS} &
$\mathbf{EMS_{\mathrm{ref}}}$ &
\textbf{Human Eval.} \\
\midrule
Human
& 91.93
& 87.14
& 90.74
& 95.31
& 94.52
& --
& --
& --
& --
& --
& -- \\

\midrule

\rowcolor{gray!15}
\multicolumn{12}{c}{\textit{Proprietary Models}} \\

\midrule

Gemini-3.1-Pro~\cite{gemini31pro_blog}
& \textbf{77.91}
& \textbf{76.57}
& \underline{69.92}
& \textbf{86.16}
& \textbf{77.87}
& \textbf{26.77}
& 38.20
& 83.02
& 26.49
& \underline{54.83}
& \underline{3.95} \\

Claude-4.6-Sonnet~\cite{anthropic2026claude46}
& 72.51
& \underline{75.43}
& 68.42
& 74.84
& 71.58
& \underline{25.81}
& \underline{38.26}
& \textbf{83.28}
& \underline{27.17}
& 54.55
& 3.89 \\

GPT-5.4~\cite{gpt54}
& 65.31
& 69.14
& 63.91
& 68.55
& 62.57
& 19.88
& 31.68
& 77.67
& 26.55
& 52.93
& 3.71 \\

\midrule

\rowcolor{gray!15}
\multicolumn{12}{c}{\textit{Open-source Models}} \\

\midrule

Kimi-k2.5~\cite{team2026kimi}
& \underline{75.87}
& 72.57
& \textbf{72.18}
& \underline{83.65}
& \underline{75.41}
& 25.77
& \textbf{41.22}
& \underline{83.09}
& \textbf{27.48}
& \textbf{54.89}
& \textbf{3.97} \\

Qwen3.5-27B~\cite{qwen3.5}
& 62.30
& 74.29
& 59.40
& 64.15
& 56.83
& 17.81
& 27.57
& 79.21
& 26.25
& 52.73
& 3.48 \\

Qwen3.5-397B-A17B~\cite{qwen3.5}
& 61.82
& 73.14
& 62.41
& 64.78
& 54.92
& 16.94
& 26.18
& 76.77
& 25.89
& 52.52
& 3.25 \\

Qwen3.5-122B-A10B~\cite{qwen3.5}
& 59.30
& 70.86
& 58.65
& 62.26
& 52.73
& 12.96
& 21.74
& 77.61
& 25.78
& 51.96
& 3.00 \\

Gemma-4-31B-It~\cite{gemma4}
& 58.82
& 64.00
& 57.89
& 60.38
& 56.01
& 16.53
& 25.75
& 78.18
& 26.26
& 52.65
& 2.95 \\

Qwen3.5-9B~\cite{qwen3.5}
& 52.82
& 55.43
& 54.89
& 61.01
& 47.27
& 8.20
& 16.07
& 73.98
& 25.45
& 50.41
& 2.73 \\

\bottomrule
\end{tabular}
}

\caption{\revise{
\textbf{Performance of evaluated MLLMs on DVBench.}
The four closed-ended dimensions are evaluated by accuracy, where
Avg. denotes their average accuracy.
For the \textit{Alignment} dimension, we report three categories:
\textbf{text-based metrics} (BLEU-2, METEOR, and BERTScore F1),
\textbf{visual grounding metrics} (EMScore and
$\mathrm{EMScore}_{\mathrm{ref}}$), and
\textbf{human evaluation}.
EMScore and $\mathrm{EMScore}_{\mathrm{ref}}$ are metrics assessing the correspondence between
generated text and visual content~\cite{shi2022emscore, wang2026howtonarrate}.
Human Eval. denotes the average rating on a five-point scale.
The Human Baseline is evaluated on a stratified 20\% subset of
the closed-ended questions.
Best and second-best results among MLLMs are highlighted in
\textbf{bold} and \underline{underline}, respectively.
}}

\label{tab:detailed_results}
\end{table*}

\subsection{Settings}
We test a total of nine MLLMs, including both open-source (Qwen3.5 series~\cite{qwen3.5}, Gemma-4~\cite{gemma4}, Kimi-k2.5~\cite{team2026kimi}) and proprietary MLLMs (Gemini-3.1-Pro~\cite{gemini31pro_blog}, Claude-4.6-Sonnet~\cite{anthropic2026claude46}, and GPT-5.4~\cite{gpt54}). Model configurations, evaluation prompts and metrics are respectively detailed in Appendices~\ref{sec:appendix_model_config}, \ref{sec:appendix_eval_prompt}, \ref{sec:appendix_evaluation_metrics}.

\revise{In the main experiment, each model follows its default frame-sampling configuration, with video-only input. To examine the influence of input configurations, we further conduct two ablations: a frame ablation with a unified 64-frame setting and different sampling rates, and a subtitle ablation comparing video-only input with video supplemented by subtitles.}

\subsection{Result}
\subsubsection{Main Result}

Table~\ref{tab:detailed_results} presents a comprehensive performance evaluation of 9 MLLMs across 5 dimensions.

\paragraph{The general superiority of proprietary models and the exceptional performance of Kimi-k2.5.} The results indicate that proprietary models generally exhibit superior performance. However, Kimi-k2.5 emerges as a notable exception, achieving the second-highest average accuracy of 75.87\% among all models, surpassing Claude-4.6-Sonnet and GPT-5.4. \revise{A similar pattern is observed in the Alignment dimension, where proprietary models and Kimi-k2.5 generally outperform the other open-source models across text-based, visual-grounding, and human evaluation metrics.}

\paragraph{The capacity of open-source models to understand data videos does not scale directly with the number of parameters.} Within the Qwen3.5 series, the relatively smaller dense model Qwen3.5-27B achieves 62.30\% average accuracy and higher Alignment performance than its larger Mixture-of-Experts counterparts, Qwen3.5-397B-A17B and Qwen3.5-122B-A10B.

\paragraph{Open-source models demonstrate parity in macro-narrative understanding but exhibit disparity in fine-grained visual perception.} Prominent open-source models display superiority in narrative comprehension. For example, Qwen3.5-27B achieves 74.29\% accuracy in the Narrative dimension, closely trailing Claude-4.6-Sonnet. However, a significant performance gap emerges within visual dimensions. Qwen3.5-27B scores only 59.40\% in Animation and 56.83\% in Chart Reasoning, lagging substantially behind proprietary models such as Gemini-3.1-Pro (69.92\% and 77.87\%) and Claude-4.6-Sonnet (68.42\% and 71.58\%).

\subsubsection{Fine-grained Analysis}
\begin{figure}[h]
    \centering
    \includegraphics[width=\linewidth]{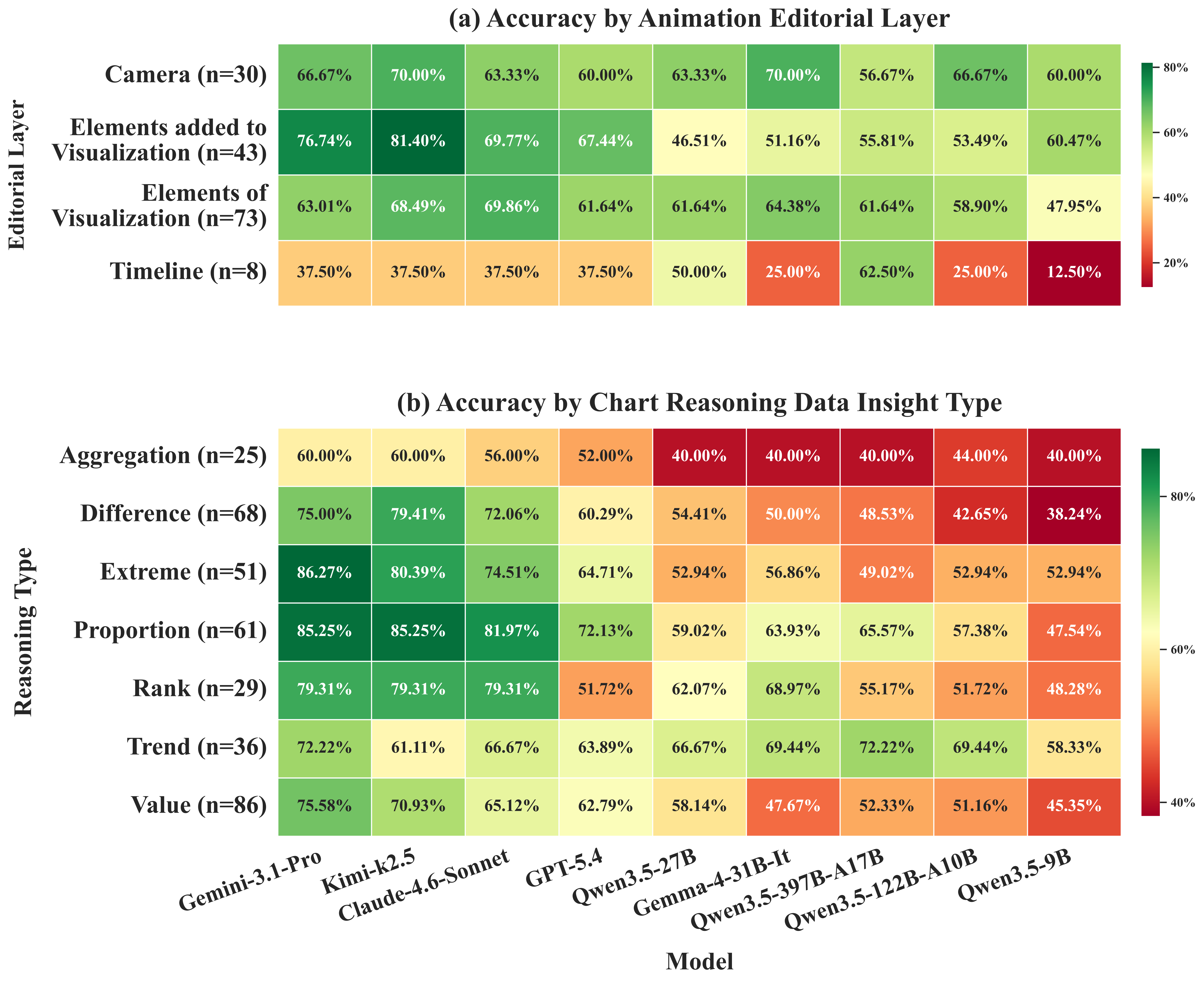}
    \caption{\textbf{Fine-grained performance across the Animation and Chart Reasoning dimensions.} (a) Accuracy distribution within the Animation task, evaluated across four editorial layers. (b) Accuracy evaluation across 10 reasoning question types. The x-axis, representing the evaluated MLLMs, is shared across both subplots.}
    \label{fig:analysis_heatmap}
\end{figure}

\begin{figure}[t] 
    \centering
    \includegraphics[width=\linewidth]{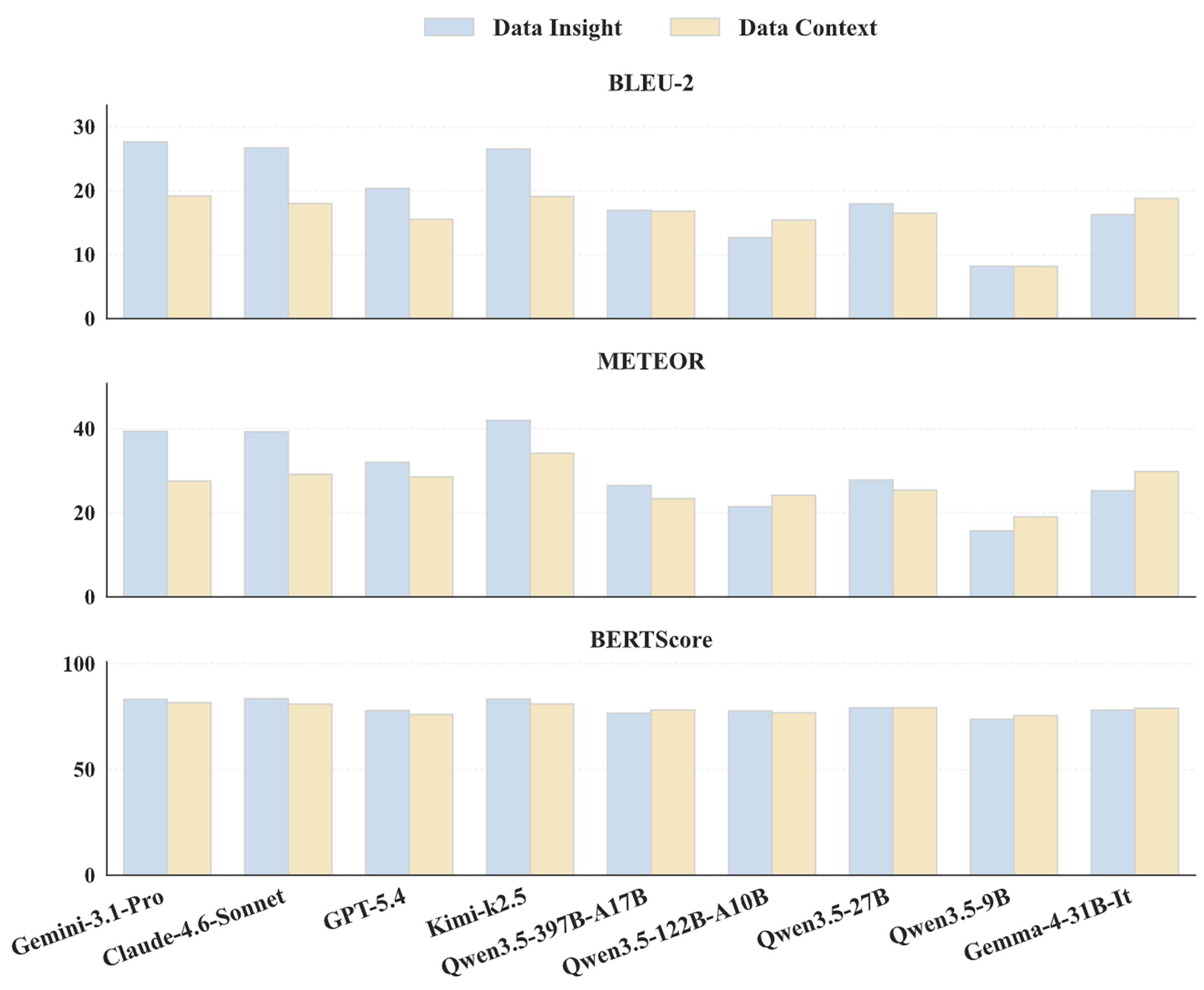}
    \caption{\textbf{Evaluation results of Alignment dimension.} We report the BERTScore, BLEU-2, and METEOR scores of 9 MLLMs on two semantic categories: \textit{Data Insight} (blue bars) and \textit{Data Context} (yellow bars). The x-axis, representing the evaluated MLLMs, is shared across all subplots.}
    \label{fig:alignment_type}
\end{figure}

\label{sec:analysis}
In this section, we further analyze model performance using the fine-grained categories defined in Sec.~\ref{sec:task_dim}. Specifically, we examine performance across the four editorial layers of Animation, the 11 data insight types of Chart Reasoning, and the two semantic categories of Alignment. We additionally investigate how video length affects model performance across different dimensions.
\begin{figure*}[!ht]
    \centering
    \includegraphics[width=\textwidth]{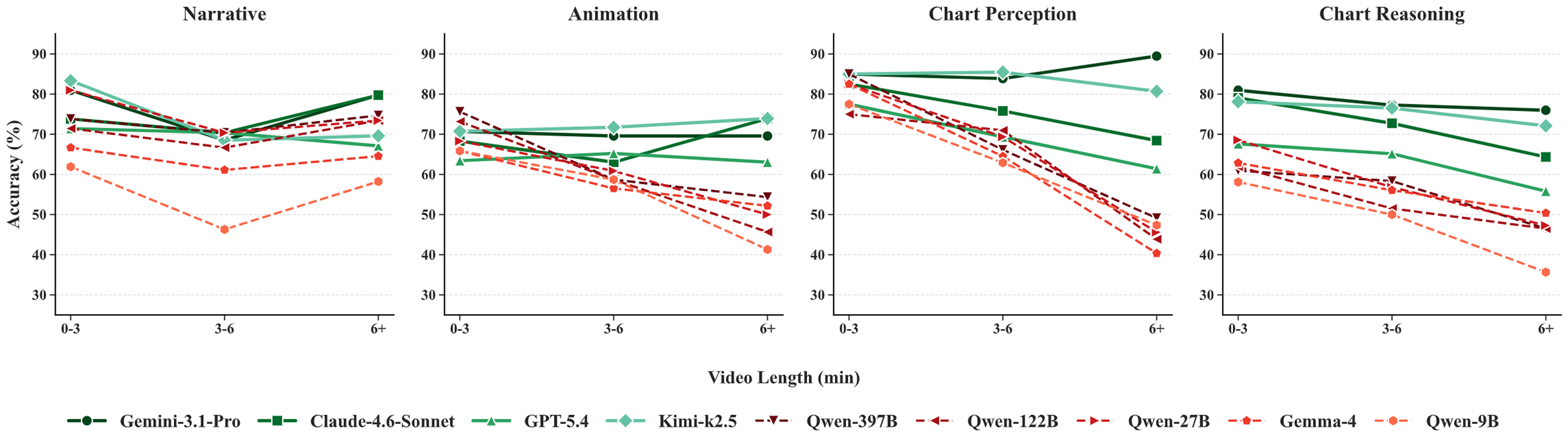}
    \caption{\textbf{Performance across varying video lengths.} The line charts illustrate the accuracy of 9 MLLMs on 4 dimensions as the video length increases from short ($\le$ 3 mins) to long ($>$ 6 mins). {\color{green}Green} lines represent strong models, while {\color{red}red} lines represent weak ones.}
    \label{fig:video_duration}
\end{figure*}

\paragraph{Animation: Failure to Capture Non-linear Animation Pacing.}
We analyze the Animation dimension from the perspective of the four editorial layers (i.e., \textit{what} is being animated) introduced in Sec.~\ref{sec:task_dim}. The four layers cover changes in visualization elements, added elements, camera perspectives, and animation pacing. Examples of the four layers are provided in Appendix~\ref{sec:appendix_case_animation}. As shown in Fig.~\ref{fig:analysis_heatmap}(a), the Timeline layer is particularly challenging, with nearly all models achieving accuracy below 37.50\%. \revise{While questions about visualization elements, added elements, and camera perspectives can often be answered by recognizing changes in visual states, Timeline questions require models to perceive how the pace of these changes evolves over time, such as acceleration and deceleration. The consistently low performance suggests that current MLLMs struggle to capture such fine-grained temporal dynamics in data videos. We further examine the effect of frame sampling on Timeline questions in Appendix~\ref{sec:appendix_timeline_sampling}. The results suggest that sparse sampling sparse sampling is one contributor rather than the cause of the poor performance.}

\begin{table*}[!ht]
\centering
\small
\setlength{\tabcolsep}{4.5pt}

\begin{minipage}[!ht]{0.49\textwidth}
\centering
\textbf{(a) Unified Frame Budget}\\[2pt]
\resizebox{\linewidth}{!}{
\begin{tabular}{lccc}
\toprule
\textbf{Model} &
\textbf{Default} &
\textbf{64F} &
\textbf{$\Delta$} \\
\midrule
Gemini-3.1-Pro
& 77.91$_{\text{1 fps}}$ & 75.15 & -2.76 \\

Claude-4.6-Sonnet
& 72.51$_{\text{100F}}$ & 69.87 & -2.64 \\

GPT-5.4
& 65.31$_{\text{64F}}$ & 65.31 & 0.00 \\

Qwen3.5-397B-A17B
& 61.82$_{\text{2 fps}}$ & 66.51 & +4.69 \\

Qwen3.5-122B-A10B
& 59.30$_{\text{2 fps}}$ & 67.73 & +8.43 \\

Qwen3.5-27B
& 62.30$_{\text{2 fps}}$ & 68.43 & +6.13 \\

Qwen3.5-9B
& 52.82$_{\text{2 fps}}$ & 60.65 & +7.83 \\
\bottomrule
\end{tabular}
}
\end{minipage}
\hfill
\begin{minipage}[!ht]{0.47\textwidth}
\centering
\textbf{(b) Effect of Sampling Rate}\\[2pt]
\resizebox{\linewidth}{!}{
\begin{tabular}{lccc}
\toprule
\textbf{Model} &
\textbf{0.5 fps} &
\textbf{1 fps} &
\textbf{2 fps} \\
\midrule
Qwen3.5-397B-A17B
& \textbf{71.07} & 70.23 & 61.82 \\

Qwen3.5-122B-A10B
& \textbf{70.73} & 69.36 & 59.30 \\

Qwen3.5-27B
& \textbf{72.14} & 70.93 & 62.30 \\

Qwen3.5-9B
& 63.47 & \textbf{64.64} & 52.82 \\
\bottomrule
\end{tabular}
}

\vspace{4pt}
\footnotesize
All values denote average accuracy over the four closed-ended dimensions.
\end{minipage}

\caption{\revise{\textbf{Ablation of frame-sampling configurations.}
(a) compares each model's default configuration with a unified
64-frame budget; subscripts indicate the default configuration.
(b) reports the effect of frame sampling rate on the Qwen3.5
models.}
}
\label{tab:frame_ablation}
\end{table*}

\begin{table}[!h]
\centering
\scriptsize
\setlength{\tabcolsep}{3pt}
\renewcommand{\arraystretch}{1.1}

\resizebox{\columnwidth}{!}{
\begin{tabular}{lrrrrr}
\toprule
\textbf{Model} &
\textbf{$\Delta$Avg.} &
\textbf{$\Delta$Narr.} &
\textbf{$\Delta$Anim.} &
\textbf{$\Delta$Perc.} &
\textbf{$\Delta$Reas.} \\
\midrule

Gemini-3.1-Pro
& \cellcolor{blue!16}+1.80
& \cellcolor{blue!33}+6.29
& \cellcolor{blue!18}+2.26
& -2.51
& \cellcolor{blue!15}+1.36 \\

Claude-4.6-Sonnet
& \cellcolor{blue!12}+0.48
& \cellcolor{blue!14}+1.14
& -0.75
& -2.51
& \cellcolor{blue!17}+1.92 \\

GPT-5.4
& \cellcolor{blue!27}+4.80
& \cellcolor{blue!35}+6.86
& \cellcolor{blue!18}+2.26
& \cellcolor{blue!17}+1.89
& \cellcolor{blue!32}+6.01 \\

Kimi-k2.5
& \cellcolor{blue!19}+2.64
& \cellcolor{blue!28}+5.14
& -2.26
& \cellcolor{blue!17}+1.88
& \cellcolor{blue!23}+3.55 \\

Qwen3.5-397B-A17B
& \cellcolor{blue!29}+5.29
& \cellcolor{blue!30}+5.72
& 0.00
& \cellcolor{blue!17}+1.89
& \cellcolor{blue!40}+8.47 \\

Qwen3.5-122B-A10B
& \cellcolor{blue!31}+5.89
& \cellcolor{blue!26}+4.57
& -0.76
& \cellcolor{blue!15}+1.26
& \cellcolor{blue!49}+10.93 \\

Qwen3.5-27B
& \cellcolor{blue!27}+4.81
& \cellcolor{blue!28}+5.14
& -0.75
& 0.00
& \cellcolor{blue!41}+8.74 \\

Qwen3.5-9B
& \cellcolor{blue!38}+7.80
& \cellcolor{blue!55}+12.57
& -2.26
& \cellcolor{blue!19}+2.51
& \cellcolor{blue!51}+11.47 \\

Gemma-4-31B-It
& \cellcolor{blue!34}+6.73
& \cellcolor{blue!51}+11.43
& \cellcolor{blue!23}+3.76
& \cellcolor{blue!15}+1.26
& \cellcolor{blue!38}+7.92 \\

\bottomrule
\end{tabular}
}
\caption{\revise{
\textbf{Subtitle Ablation.}
Absolute performance changes between video-only input and video input supplemented with subtitles. Positive values indicate performance gains with subtitle input, with darker blue backgrounds denoting larger improvements; non-positive values are left unshaded. The corresponding video-only results are reported in Table~\ref{tab:detailed_results}.
}}
\label{tab:subtitle_ablation}
\end{table}

\paragraph{Chart Reasoning: Dynamic Charts Induce Severe Counting Hallucinations.} 
Within the Chart Reasoning dimension, Aggregation evaluates the ability to count and synthesize information across visual elements. These questions require models to identify the elements that satisfy a given condition and determine their exact number, sometimes by tracking the relevant elements across multiple frames. As shown in Fig.~\ref{fig:analysis_heatmap}(b), Aggregation is one of the most challenging reasoning categories, with several open-source models achieving accuracy around 40.00\%. This suggests that current MLLMs still struggle with fine-grained counting of visual elements and condition-based aggregation in dynamic charts. Detailed cases are provided in Appendix~\ref{sec:appendix_case_reasoning}.

\paragraph{Alignment: The Divergence in Insight and Context Grounding.}
\revise{Text-based metrics, visual grounding metrics, and human evaluation reveal a similar pattern: Kimi-k2.5, Gemini-3.1-Pro, and Claude-4.6-Sonnet achieve the strongest performance. However, we observe a clear divergence in the text-based metrics. As shown in Fig~\ref{fig:alignment_type}, BLEU-2 and METEOR show noticeable differences between the two categories: proprietary models and Kimi-k2.5 generally perform better on \textit{Data Insight} than on \textit{Data Context}, whereas the other open-source models exhibit a larger drop on \textit{Data Insight}. Error analysis further shows that this gap is mainly associated with visual-text grounding errors rather than phrasing differences. Detailed analysis and examples are provided in Appendix~\ref{sec:appendix_case_alignment}.}

\paragraph{Video Length: Strong Models Are Not Susceptible to Long Videos.} We analyze model performance across short, medium, and long videos. As illustrated in Fig~\ref{fig:video_duration}, leading proprietary models and Kimi-k2.5 demonstrate remarkable temporal resilience; their overall performance plateaus even as the video length extends beyond 6 minutes. In contrast, weaker models experience a nearly linear decline in accuracy as the video length increases. Furthermore, we observe that the impact of video length is task-dependent. For high-level Narrative dimension, which requires holistic semantic abstraction, the performance of nearly all models remains relatively stable, unaffected by increased video length, except for Qwen3.5-9B. However, in fine-grained visual tasks (illustrated in the three rightmost subfigures), weaker models struggle with long videos, with accuracy frequently plummeting below 50\%.

\revise{\subsubsection{Ablation Studies}}
\revise{\paragraph{Frame Ablation: More Frames Do Not Necessarily Help.}
We further evaluate all applicable models using 64 uniformly sampled frames, a common setting within their supported input ranges. As shown in Table~\ref{tab:frame_ablation}(a), Gemini-3.1-Pro and Claude-4.6-Sonnet remain the strongest models. Within the Qwen3.5 family, Qwen3.5-27B continues to outperform the larger models. These results show that the model ranking under a unified frame budget remains consistent with that observed in the main experiment, while the performance gaps between models become smaller.}

\revise{
We further evaluate the Qwen3.5 models under different frame sampling rates, as shown in Table~\ref{tab:frame_ablation}(b). Performance does not improve monotonically with denser sampling: the 2 fps setting consistently underperforms 0.5 and 1 fps across all Qwen3.5 models, while 0.5 or 1 fps achieves the best performance. This indicates that increasing the number of sampled frames does not necessarily provide more useful evidence for understanding data videos.
}

\revise{\paragraph{Subtitle Input: Subtitles Primarily Benefit High-Level Semantic Understanding.}
We compare video-only input with video input supplemented by subtitles across all nine models. As shown in Table~\ref{tab:subtitle_ablation}, subtitles improve the average performance of all models, with larger gains for the Qwen3.5 and Gemma models than for most proprietary models. The improvements are concentrated in Narrative and Chart Reasoning. In particular, Chart Reasoning improves by 7.92--11.47\% for the Qwen3.5 and Gemma models. In contrast, gains on Chart Perception and Animation are smaller and less consistent. Overall, subtitle input provides greater benefits for semantic understanding than for low-level visual perception.}

\section{Conclusion}
\label{sec:bibtex}
We introduce DVBench, the first comprehensive benchmark evaluating MLLMs on data video understanding. We divide the data video understanding task into 5 dimensions. We curate a dataset of 300 videos and 1,000 QA pairs via a semi-automatic pipeline and expert verification. Evaluation of 9 MLLMs demonstrates the superiority of proprietary models and exposes unexpected results in parameter scaling and cross-dimensional proficiency. Fine-grained analyses reveal weaknesses across different evaluation dimensions, while ablations demonstrate the impact of different input configurations. These findings provide valuable insights for future research.

\section*{Limitations}
While our semi-automatic construction pipeline combined with expert validation
ensures high data fidelity, the reliance on human review currently limits the
scale of DVBench. \revise{We also observe that performance on the Chart
Perception dimension is relatively saturated among frontier models, suggesting
that future benchmark iterations should incorporate more challenging
fine-grained perception tasks to better distinguish model capabilities.}
Future work could further expand the benchmark to include multilingual data
videos, 3D visualizations, and AR infographics. Moreover, as native
end-to-end video models advance toward processing denser audio-visual streams,
they may provide stronger capabilities for the fine-grained temporal tasks
that remain challenging for current MLLMs. DVBench's extensible taxonomy is
designed to accommodate increasingly complex reasoning tasks and support the
continued evaluation of future model architectures.

\section*{Acknowledgments}
We want to thank the reviewers for their suggestions. This work was supported by the National Natural Science Foundation of China (No. 62472099), the AI for Science Program of the Shanghai Municipal Commission of Economy and Informatization (Grant No. 2025-GZL-RGZN-BTBX-02028), the Fundamental and Interdisciplinary Disciplines Breakthrough Plan of the Ministry of Education of China (No. JYB2025XDXM904), and the Ji Hua Laboratory S\&T Program (No. X250881UG250).

\bibliography{custom}

\clearpage
\appendix
\section*{Appendix}

\section{Potential Risks}
\label{sec:appendix_risks}
While DVBench is designed strictly for the academic evaluation of MLLMs, the use of data videos for model development and evaluation may still involve potential downstream risks. Data videos often present real-world statistics and analytical narratives concerning socio-economic, geopolitical, environmental, and public health topics, and may therefore contain sensitive information or reflect particular perspectives.

The potential risks are relatively constrained in DVBench because all videos are curated from previously published academic works rather than unrestricted online media. In addition, the dataset curation process includes manual review and quality control to identify and remove problematic or misleading content where possible. These measures reduce, but do not completely eliminate, the potential risks associated with the dataset.

First, there is a risk of bias inheritance. Although the videos are sourced from established prior works, their narratives may still contain implicit assumptions, framing choices, or subjective interpretations. Models trained or evaluated on such content may consequently reproduce or amplify these biases.

Second, there is a risk of misinformation generation. Our error analysis shows that current open-source MLLMs frequently exhibit numerical hallucinations and weak visual grounding when interpreting charts and animated visualizations. If such limitations are not properly mitigated, deploying these models in applications such as automated journalism, educational analytics, or data storytelling could lead to fabricated insights or visually persuasive but factually incorrect narratives.


\section{Dataset Construction Details}
\subsection{Licenses and Intended Use}
\label{sec:appendix_license}
DVBench is developed exclusively for non-commercial academic research and evaluation of MLLMs. The benchmark annotations, extracted structured information, QA pairs, and accompanying metadata are released under the \textbf{CC BY-NC-SA 4.0} license.

The benchmark contains 300 data videos collected from previously published research works~\cite{yang2021freytagPyramid, shi2021communicating, cheng2022investigating, xu2022wow, xu2023end, gao2025sceneloom, gunturu2025mapstory}. These videos were originally made publicly available on platforms such as YouTube and Vimeo. We do not claim ownership of the original video, audio, or visual materials, whose copyrights remain with their respective creators and copyright holders. Accordingly, DVBench is intended only to facilitate standardized academic evaluation, and users are responsible for complying with applicable copyright laws and the terms of the original hosting platforms.

To respect the rights of content owners, we provide a removal mechanism for third-party content included in the benchmark. If a copyright holder believes that specific content should not be included, we will promptly review and address reasonable removal requests.

During dataset construction and manual quality control, the videos and QA pairs were also reviewed to minimize personally identifiable information (PII), offensive language, and potentially harmful content. However, given that the source videos were originally created by third parties, we cannot guarantee the complete absence of such content.

\subsection{Dataset Construction Prompt}
\label{sec:appendix_dataset_prompt}
In this section, we provide the detailed prompt templates utilized in our semi-automatic benchmark construction pipeline.

\begin{tcolorbox}[
    title={Prompt for Narrative QA Generation},
    fonttitle=\small\bfseries,
    colbacktitle=gray!70!white,
    coltitle=white,
    colback=gray!5!white,
    colframe=gray!70!white,
    arc=2mm,
    boxrule=0.5pt,
    breakable,
    left=6pt, right=6pt, top=6pt, bottom=6pt
]
\small
\textbf{[Role]} \\
You are an expert in Data Visualization Analysis and Video Understanding. 

\vspace{0.5em}
\textbf{[Task Category]} \\
Your task is to analyze the provided video content to generate QA pairs evaluating the viewer's narrative comprehension. The actual video title is: \texttt{\{title\}}. Your generated questions must strictly fall into one of two categories:

\vspace{0.5em}
\textbf{Category 1: Narrative Structure} \\
Design questions focusing on the storytelling logic and visual-narrative mapping.
\begin{itemize}
    \item \textbf{Type 1: VIS $\rightarrow$ Topic/Intent.} Ask about a specific visualization and require its narrative intent. \textit{Distractor Requirements:} (1) Metric-focused (raw data insight); (2) Logical but Irrelevant; (3) Extraneous Insight. 
    \item \textbf{Type 2: Topic/Intent $\rightarrow$ VIS.} Provide a narrative purpose and ask which visual method conveys it. \textit{Distractor Requirements:} Visualizations that appear in the video but do not serve this specific purpose.
    \item \textbf{Type 3: Logical.} Inquire about the logical connection between two chart transitions (e.g., "The bar chart supports the line chart") or infer the reason for the next visual content.
\end{itemize}

\vspace{0.5em}
\textbf{Category 2: Information Synopsis} \\
Design questions focusing on the theme and subjective tone.
\begin{itemize}
    \item \textbf{Title:} What is the best title or main topic of the video? (Use the provided title to guide the answer, and design plausible alternative titles as distractors).
    \item \textbf{Attitude:} What is the attitude of the video towards the main topic? (Output a single adjective like optimistic, pessimistic, or neutral).
\end{itemize}

\vspace{0.5em}
\textbf{[Output Format]}
\vspace{0.3em}
\hrule
\vspace{0.3em}
\begin{flushleft}
\ttfamily
[\par
\quad \{\par
\quad\quad "category": "Narrative Structure",\par
\quad\quad "type": "[VIS -> Topic, Topic -> VIS, or Logical]",\par
\quad\quad "question": "[Insert specific question]",\par
\quad\quad "answer": "[Insert specific answer]",\par
\quad\quad "distractors": ["[Distractor 1]", "[Distractor 2]", "[Distractor 3]"],\par
\quad\quad "evidence\_timestamp": "[Start\_Time - End\_Time]"\par
\quad \},\par
\quad \{\par
\quad\quad "category": "Information Synopsis",\par
\quad\quad "type": "[Title or Attitude]",\par
\quad\quad "question": "[Insert specific question]",\par
\quad\quad "answer": "[Insert specific answer]",\par
\quad\quad "distractors": ["[Distractor 1]", "[Distractor 2]", "[Distractor 3]"],\par
\quad\quad "evidence\_timestamp": "[Start\_Time - End\_Time]"\par
\quad \}\par
]
\end{flushleft}
\end{tcolorbox}
\vspace{1em}

\begin{tcolorbox}[
    title={Prompt for Data Clip and Data Insight Extraction}, 
    fonttitle=\small\bfseries,  
    colbacktitle=gray!70!white, 
    coltitle=white,             
    colback=gray!5!white,       
    colframe=gray!70!white,     
    arc=2mm,              
    boxrule=0.5pt,        
    breakable,            
    left=6pt, right=6pt, top=6pt, bottom=6pt 
]
\small
\textbf{[Role]} \\
You are an expert in Data Visualization Analysis and Video Understanding.

\vspace{0.5em}
\textbf{[Task]} \\
 Your task is to analyze the provided video, identify video clips that includes chart, and identify data insight for each chart, such as values (e.g., the death number in 2010 was 106), trends (e.g., the birth rate is increasing from 2000 to 2005), or ranks (e.g., China's military ranks 1st).

\vspace{0.5em}
\textbf{[Constraints]}
\begin{enumerate}
    \item \textbf{Visual Grounding:} Every answer must be derived solely from the video's visual content. Avoid using external prior knowledge.
    \item \textbf{Objectivity:} Questions must have a single, unambiguous, and deterministic answer.
    \item \textbf{Conciseness:} Answers should be brief (a single word, a number, or a short phrase).
    \item \textbf{Chart Clustering:} Organize the generated QA pairs into groups based on the specific chart or visual component they describe. Each group should represent a single, distinct chart (e.g., "World Population Line Chart" or "Income Distribution Map"). Ensure that all data points, titles, and axis information for a specific chart are clustered together in the output.
\end{enumerate}

\vspace{0.5em}
\textbf{[Output Format]} \\
\vspace{0.3em}
\hrule
\vspace{0.3em}
\begin{flushleft}
\ttfamily
[\par
\quad \{\par
\quad\quad "chart\_group": "[Insert name of the specific chart]",\par
\quad\quad "insights": [\par
\quad\quad\quad \{\par
\quad\quad\quad\quad "description": "A description of the data insight",\par
\quad\quad\quad\quad "evidence\_timestamp": "[Start\_Time - End\_Time]"\par
\quad\quad\quad \}\par
\quad\quad ]\par
\quad \}\par
]
\end{flushleft}
\end{tcolorbox}
\vspace{1em}

\begin{tcolorbox}[
    title={Prompt for Data Insight Extraction from Subtitle}, 
    fonttitle=\small\bfseries,
    colbacktitle=gray!70!white,
    coltitle=white,
    colback=gray!5!white,
    colframe=gray!70!white,
    arc=2mm,
    boxrule=0.5pt,
    breakable,
    left=6pt, right=6pt, top=6pt, bottom=6pt
]
\small
\textbf{[Role]} \\
You are an expert in Data Visualization Analysis and Video Understanding. 

\vspace{0.5em}
\textbf{[Task]} \\
Extract data insight from the subtitle of data video.

\vspace{0.5em}
\textbf{[Extraction Scope]}
\begin{enumerate}
    \item \textbf{Precise Data:} Extract explicitly mentioned percentages, monetary amounts, population figures, and other hard numbers.
    \item \textbf{Qualitative Descriptions:} Even in the absence of specific numbers, you must extract descriptions of trends (growth/decline/stagnation), intensity (sharp/gradual), comparisons (surpassing/lagging), or structural changes (bridging the gap/moving into the middle class).
\end{enumerate}

\vspace{0.5em}
\textbf{[Insight Definition]}
\begin{itemize}
    \item \textbf{description:} A natural language description of the data insight.
    \item \textbf{type:} Select the most appropriate data insight category listed below.
    \item \textbf{time\_stamp:} The period when the insight is mentioned in the video time stamp. From when the complete sentence begins to when the sentence ends (e.g., 00:00:14,400 $\rightarrow$ 00:00:18,000).
\end{itemize}

\vspace{0.5em}
\textbf{[Insight Type Definition]}
\begin{itemize}
    \item \textbf{Value:} Retrieve exact values under specific criteria.
    \item \textbf{Proportion:} Measure the percentage of selected attributes within a set.
    \item \textbf{Difference:} Compare attributes or temporal changes.
    \item \textbf{Distribution:} Show the breakdown or share across attributes.
    \item \textbf{Trend:} Present a general tendency over time.
    \item \textbf{Rank:} Sort attributes based on their values.
    \item \textbf{Aggregation:} Calculate statistical indicators like average, sum, or count.
    \item \textbf{Association:} Identify correlations between multiple attributes.
    \item \textbf{Extreme:} Find the top/bottom cases or "-est" values.
    \item \textbf{Categorization:} Select attributes satisfying specific conditions.
\end{itemize}

\vspace{0.5em}
\textbf{[Output Format]}
\vspace{0.3em}
\hrule
\vspace{0.3em}
\begin{flushleft}
\ttfamily
\{\par
\quad "description": "[Insert natural language description]",\par
\quad "type": "[Insert one of the 10 fact types]",\par
\quad "time\_stamp": "[Start\_Time $\rightarrow$ End\_Time]"\par
\}\par

\vspace{0.5em}
\textbf{[Subtitle]}\\
\textcolor{blue}{\{subtitle\}}
\end{flushleft}
\end{tcolorbox}
\vspace{1em}

\begin{tcolorbox}[
    title={Prompt for Insight Formalization},
    fonttitle=\small\bfseries,
    colbacktitle=gray!70!white,
    coltitle=white,
    colback=gray!5!white,
    colframe=gray!70!white,
    arc=2mm,
    boxrule=0.5pt,
    breakable,
    left=6pt, right=6pt, top=6pt, bottom=6pt
]
\small
\textbf{[Role]} \\
You are a data extraction expert. Your task is to convert a list of natural language descriptions of data facts into a structured Data Fact format.

\vspace{0.5em}
\textbf{[Fact Definition]} \\
Each Fact consists of the following components:
\begin{itemize}
    \item \textbf{description:} A natural language description of the fact.
    \item \textbf{type:} Select the most appropriate fact category listed below.
    \item \textbf{parameters:} Specific descriptive arguments for the fact type.
    \item \textbf{measure(s):} Numerical fields treated as dependent variables (e.g., Sales, Units, Price), derived from functions like SUM, AVG, or COUNT. Use original words from the subtitles as much as possible. Use complete phrases (e.g., "death prevented" rather than "death").
    \item \textbf{subject:} Defines the scope of the fact through three factors:
    \begin{itemize}
        \item \textit{context:} The data subspace defined strictly by filters applied to specific table columns (dimensions). This field must only contain direct column-value filters found in the data table. Do not include narrative context, assumptions, or descriptive phrases.
        \item \textit{breakdown(s):} Dimensions used to divide the subspace into groups.
        \item \textit{focus:} Specific groups within the subspace to be highlighted or emphasized.
    \end{itemize}
    \item \textbf{time\_stamp:} The period when the fact is mentioned in the video time stamp. From when the complete sentence begins to when the sentence ends (e.g., 00:00:14,400 $\rightarrow$ 00:00:18,000).
\end{itemize}

\vspace{0.5em}
\textbf{[Fact Type Definition]}
\begin{itemize}
    \item \textbf{Value:} Retrieve exact values under specific criteria.
    \item \textbf{Proportion:} Measure the percentage of selected attributes within a set.
    \item \textbf{Difference:} Compare attributes or temporal changes.
    \item \textbf{Distribution:} Show the breakdown or share across attributes.
    \item \textbf{Trend:} Present a general tendency over time.
    \item \textbf{Rank:} Sort attributes based on their values.
    \item \textbf{Aggregation:} Calculate statistical indicators like average, sum, or count.
    \item \textbf{Association:} Identify correlations between multiple attributes.
    \item \textbf{Extreme:} Find the top/bottom cases or "-est" values.
    \item \textbf{Categorization:} Select attributes satisfying specific conditions.
    \item \textbf{Outlier:} Identify unexpected or statistically abnormal data points.
\end{itemize}

\vspace{0.5em}
\textbf{[Output Format \& Examples]} \\
\vspace{0.3em}
\hrule
\vspace{0.3em}
\begin{flushleft}
\ttfamily
[\par
\quad \{\par
\quad\quad "description": "Protein takes 66\% in the diet on Sunday.",\par
\quad\quad "type": "Proportion",\par
\quad\quad "parameters": "66\%",\par
\quad\quad "measure": "Dietary Intake",\par
\quad\quad "subject": \{\par
\quad\quad\quad "context": ["Sunday"],\par
\quad\quad\quad "breakdown": "Nutrient",\par
\quad\quad\quad "focus": ["Protein"]\par
\quad\quad \}\par
\quad \},\par
\quad \{\par
\quad\quad "description": "The Sales of iOS increased over the years.",\par
\quad\quad "type": "Trend",\par
\quad\quad "parameters": "increase",\par
\quad\quad "measure": "Sales",\par
\quad\quad "subject": \{\par
\quad\quad\quad "context": ["iOS"],\par
\quad\quad\quad "breakdown": "Year",\par
\quad\quad\quad "focus": null\par
\quad\quad \}\par
\quad \},\par
\quad \{\par
\quad\quad "description": "The national average price for regular gas was \$4.06 in July 2008.",\par
\quad\quad "type": "Aggregation",\par
\quad\quad "parameters": "4.06",\par
\quad\quad "measure": "AVG(national price)",\par
\quad\quad "subject": \{\par
\quad\quad\quad "context": ["2008", "July"],\par
\quad\quad\quad "breakdown": null,\par
\quad\quad\quad "focus": null\par
\quad\quad \}\par
\quad \},\par
\quad ...
]\par

\vspace{0.8em}
\textbf{[Input Description]} \\
\textcolor{blue}{\{description\}}
\end{flushleft}
\end{tcolorbox}
\vspace{1em}

\begin{tcolorbox}[
    title={Prompt for Cross-Frame Reasoning QA Generation},
    fonttitle=\small\bfseries,
    colbacktitle=gray!70!white,
    coltitle=white,
    colback=gray!5!white,
    colframe=gray!70!white,
    arc=2mm,
    boxrule=0.5pt,
    breakable,
    left=6pt, right=6pt, top=6pt, bottom=6pt
]
\small
\textbf{[Role]} \\
You are a data extraction and QA generation expert. Your task is to generate complex cross-frame reasoning questions for a data video based on a provided collection of structured Data Facts.

\vspace{0.5em}
\textbf{[Generation Strategies]} \\
You will receive Data Facts grouped either by \textbf{Measure} or by \textbf{Subject}. Apply the corresponding generation logic based on the input type:

\begin{itemize}
    \item \textbf{Strategy 1: Grouped by Measure (Cross-Attribute Derivation)} 
    \begin{itemize}
        \item \textit{Condition:} Facts share the same Measure (dependent variable) but differ in their Subject attributes (e.g., Temporal, Spatial, or Categorical differences).
        \item \textit{Logic:} Generate questions requiring mathematical calculations (differences, multiples, ratios) or logical comparisons (ranking changes, share displacement) based on facts distributed across different video timestamps.
    \end{itemize}
    \item \textbf{Strategy 2: Grouped by Subject (Point-to-Surface Query)}
    \begin{itemize}
        \item \textit{Condition:} Facts share a similar Subject (e.g., the same entity or year) but track different Measures across different timestamps.
        \item \textit{Logic:} Identify a specific context/event implicit in Fact 1 to use as a positioning anchor. Without explicitly stating this background, ask to extract the numerical value or description from Fact 2 that corresponds to the same background (e.g., "When [Event in Fact 1] occurred, what was the [Measure in Fact 2]?").
    \end{itemize}
\end{itemize}

\vspace{0.5em}
\textbf{[Constraints]}
\begin{enumerate}
    \item \textbf{Mandatory Cross-Frame:} Questions must link $\ge 2$ facts from different timestamps.
    \item \textbf{Evidence Traceability:} The output must cite the specific atomic fact descriptions and their original timestamps.
    \item \textbf{Conciseness and Directness:} Questions must be clear and focused. Answers must only contain the data fact parameters, such as specific numerical values, a delta, or a status (e.g., "50\%", "200 tons", "increase"). Do not use full sentences.
\end{enumerate}

\vspace{0.5em}
\textbf{[Output Format \& Examples (JSON)]}
\vspace{0.3em}
\hrule
\vspace{0.3em}
\begin{flushleft}
\ttfamily
[\par
\quad \textit{// Example for Strategy 1: Grouped by Measure}\par
\quad \{\par
\quad\quad "question": "Between 1500 and 1700, by how many percentage points did the percentage of death rate in the world decrease?",\par
\quad\quad "answer": "50\%",\par
\quad\quad "evidence": [\par
\quad\quad\quad \{"fact": "In 1800, the percentage of death rate in the world was 85\%.", "time\_stamp": "00:00:27 $\rightarrow$ 00:00:38"\},\par
\quad\quad\quad \{"fact": "In 1990, the percentage of death rate in the world was 35\%.", "time\_stamp": "00:01:23 $\rightarrow$ 00:01:30"\}\par
\quad\quad ],\par
\quad\quad "reasoning": "Calculates the difference between the death rate in the two timestamps."\par
\quad \},\par
\quad \textit{// Example for Strategy 2: Grouped by Subject}\par
\quad \{\par
\quad\quad "question": "At the point when Starship's inaugural flight generated 17 million pounds of thrust, what was its specified payload capacity to low Earth orbit?",\par
\quad\quad "answer": "150 metric tons",\par
\quad\quad "evidence": [\par
\quad\quad\quad \{"fact": "Starship's first flight generated a record 17 million pounds of thrust.", "time\_stamp": "00:05:12 $\rightarrow$ 00:05:18"\},\par
\quad\quad\quad \{"fact": "The Starship is designed to carry 150 metric tons to low Earth orbit.", "time\_stamp": "00:12:45 $\rightarrow$ 00:12:52"\}\par
\quad\quad ],\par
\quad\quad "reasoning": "Fact 1 anchors the context (thrust event). Fact 2 provides the payload measure for that same subject."\par
\quad \}\par
]\par
\vspace{0.8em}
\textbf{[Input Fact Clusters]} \
\textcolor{blue}{\{fact\_grouped\_by\_measure\_or\_subject\}}
\end{flushleft}
\end{tcolorbox}
\vspace{1em}

\subsection{Annotation Details}
\label{sec:appendix_annotation_interface}

\paragraph{Annotator Background.}
The benchmark annotation, fact-checking, and QA verification were conducted by two authors with expertise in data visualization and prior experience in creating and analyzing data videos. One annotator holds a Master's degree, and the other is a PhD candidate.

\revise{\paragraph{Question Selection Criteria.}
Before constructing the full benchmark, the two annotators jointly reviewed 10 data videos to establish consistent criteria for question selection. During this calibration stage, they compiled an initial pool of candidate questions from three sources: automatically extracted data insights, questions independently proposed by the annotators, and human-revised or synthesized insights based on the generated candidates. Semantically similar questions were merged and duplicates were removed, resulting in a pool of 88 candidate questions.}

\revise{The two annotators then independently selected questions from this pool for inclusion in the benchmark, achieving a raw agreement rate of 94.32\% and a Cohen's $\kappa$ of 0.88. All disagreements were subsequently discussed and resolved. Based on this calibration process, the final selection criteria required each question to: (1) target a core insight conveyed by the video rather than an incidental detail; (2) be grounded in the original visual or narrative evidence; and (3) require sufficient understanding or reasoning, such as integrating information across multiple scene transitions, tracking dynamically changing chart elements, or performing non-trivial data interpretation.}

\revise{\paragraph{QA Curation and Verification.}
Following the established protocol, one annotator constructed the complete QA set, including questions, answers, and distractors. The second annotator independently inspected 10\% of the resulting QA pairs, checking question validity, answer correctness, evidence grounding, and consistency with the assigned evaluation dimension. Potential ambiguities or inconsistencies were resolved through discussion before finalizing the benchmark.}

\begin{figure*}[htbp]
    \centering
    \includegraphics[width=\linewidth]{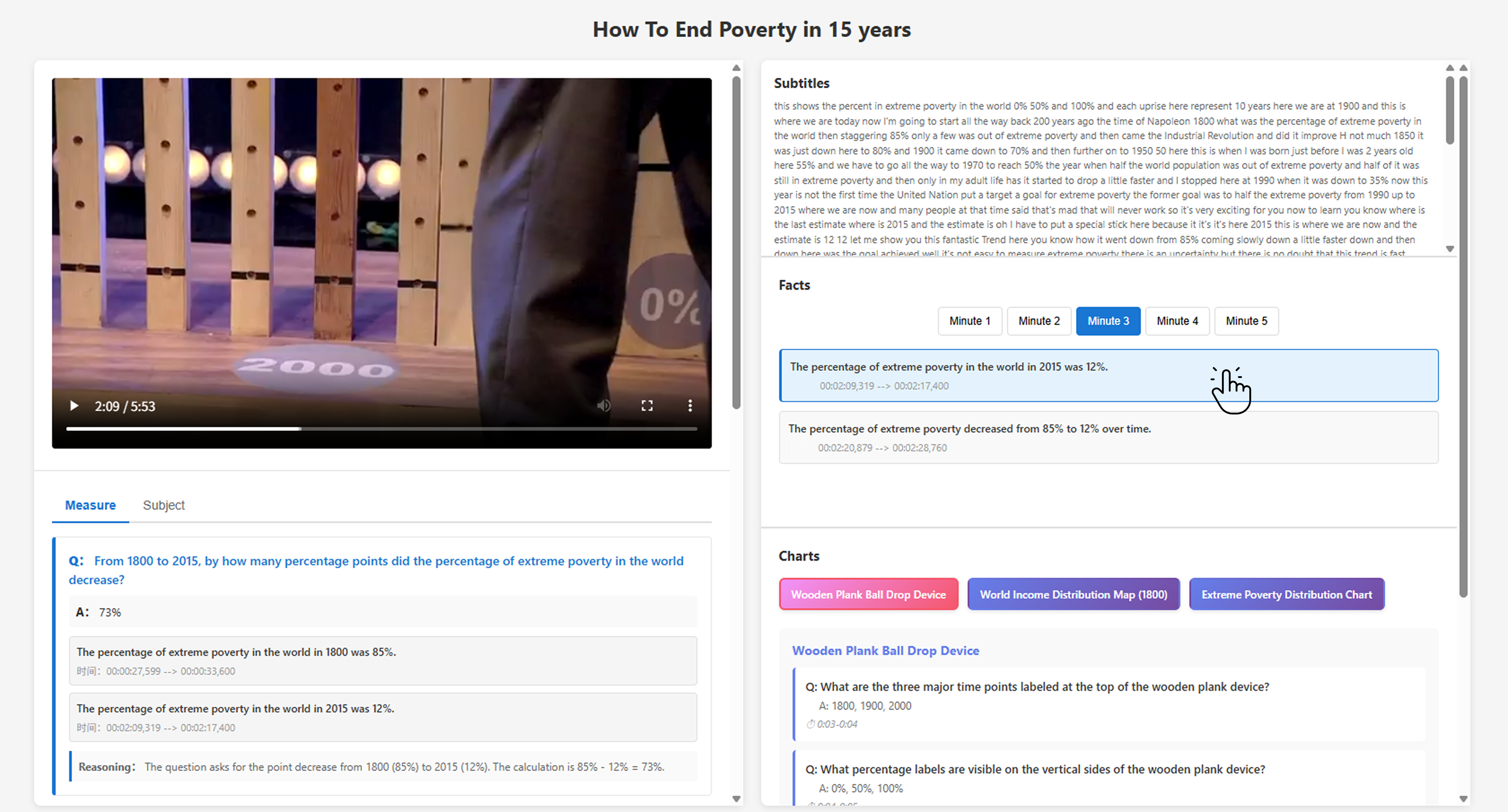}
    \caption{\textbf{Annotation interface of DVBench.} The layout is optimized for efficient verification. The left panel contains synchronized video playback (top) and cross-frame reasoning questions with their reasoning processes (bottom). The right panel displays the video subtitles (top), extracted data insights (middle), and identified data clips alongside chart perception QA pairs (bottom). An interactive click-to-seek function allows annotators to click on any insights or questions to automatically jump to the relevant timestamp for rapid visual verification.}
    \label{fig:app_ui}
\end{figure*}

\paragraph{Annotation Interface.}
To facilitate the human-in-the-loop annotation and verification process, we developed a web-based annotation interface, as illustrated in Fig.~\ref{fig:app_ui}. The interface integrates video playback, subtitles, extracted data insights, identified data clips, generated QA candidates, and cross-frame reasoning information within a unified workspace. The left panel contains a synchronized video player at the top and displays cross-frame reasoning questions, grouped by their measures and subjects, together with the corresponding reasoning information at the bottom. The right panel contains three vertically organized sections showing the subtitles, extracted data insights, and identified data clips with QA candidates, respectively. The interface further supports interactive temporal navigation across different annotation components. By clicking an extracted data fact, data clip, or question, annotators can directly navigate the video to its corresponding timestamp. This synchronization enables annotators to efficiently verify the relationship among visual evidence, subtitles, extracted insights, and QA candidates. Annotators can then revise candidate content, verify answers and distractors, and assign fine-grained categories according to the evaluation taxonomy.


\section{More Dataset Statistics}
\label{sec:appendix_dataset_statistics}
Table~\ref{tab:dvbench_task_stats} provides a detailed statistics of question and option lengths across four core evaluation dimensions. Table~\ref{tab:dvbench_alignment_stats} shows the length of the collected video subtitles and the length of missing subtitles in subtitle cloze tasks.

\begin{table}[htbp]
  \centering
  \small
  \renewcommand{\arraystretch}{1.2}
  \resizebox{0.9\columnwidth}{!}{
    \begin{tabular}{l c c}
      \toprule
      \textbf{Dimension} & \textbf{Question Len.} & \textbf{Option Len.} \\
      \midrule
      Narrative        & 10.5 (6 -- 29) & 7.4 (1 -- 30) \\
      Animation        & 12.4 (7 -- 26) & 6.4 (1 -- 21) \\
      Chart Perception & 12.2 (5 -- 24) & 2.5 (1 -- 12) \\
      Chart Reasoning  & 14.3 (3 -- 32) & 1.3 (1 -- 11) \\     
      \bottomrule
    \end{tabular}
  }
  \caption{Question and option length statistics. Values are presented as \textit{Mean (Range: Min--Max)}.}
  \label{tab:dvbench_task_stats}
\end{table}

\begin{table}[htbp]
  \centering
  \small
  \renewcommand{\arraystretch}{1.2}
  \resizebox{0.9\columnwidth}{!}{
    \begin{tabular}{l c c c}
      \toprule
      \textbf{Alignment} & \textbf{Max Length} & \textbf{Min Length} & \textbf{Avg Length} \\
      \midrule
      \textit{Video Subtitle} & 12035 & 5 & 1190.0 \\
      \textit{Answer} & 47 & 4 & 15.4 \\
      \bottomrule
    \end{tabular}
  }
    \caption{Statistics on the lengths of full video subtitles and the length of missing subtitles that need models to predict.}
  \label{tab:dvbench_alignment_stats}
\end{table}


\section{Evaluation Details}
\subsection{Model Configurations}
\label{sec:appendix_model_config}
Table~\ref{tab:model_config} summarizes the model versions, visual input configurations, decoding hyperparameters, and inference backends used in our evaluation. To ensure deterministic generation and maximum reproducibility, we set the \texttt{temperature} to 0 and \texttt{top-p} to 1 wherever applicable. For locally deployed open-source models running via vLLM, we explicitly enforce \texttt{do\_sample=False}.

In main experiment, we adopt a default sampling rate for Gemini (1 fps), Gemma (1 fps), and Qwen series (2 fps) models. For GPT-5.4 and Claude-4.6-Sonnet, taking into account the strict payload size limitations of their API requests, we uniformly sample a fixed number of frames per video (64 frames for GPT-5.4 and 100 frames for Claude-4.6-Sonnet).
\begin{table*}[htbp]
  \centering
  \small
  \renewcommand{\arraystretch}{1.2} 
  \resizebox{\textwidth}{!}{ 
    \begin{tabular}{l l c c c c l}
      \toprule
      \textbf{Model} & \textbf{Version} & \textbf{Input Frames} & \textbf{\texttt{do\_sample}} & \textbf{\texttt{temp}} & \textbf{\texttt{top-p}} & \textbf{Inference} \\
      \midrule
      \rowcolor{gray!15} 
      \multicolumn{7}{c}{\textit{Proprietary}} \\
      \midrule
      GPT-5.4 & \texttt{gpt-5.4} & 64 & $-$ & 0 & 1 & API \\
      Gemini-3.1-Pro & \texttt{gemini-3.1-pro-preview} & 1 fps & $-$ & 0 & 1 & API \\
      Claude-4.6-Sonnet & \texttt{claude-sonnet-4-6} & 100 & $-$ & 0 & $-$ & API \\
      
      \midrule
      \rowcolor{gray!15} 
      \multicolumn{7}{c}{\textit{Open-Source Models}} \\
      \midrule
      Kimi-k2.5 & \texttt{moonshotai/Kimi-K2.5} & $-$ & $-$ & $-$ & $-$ & API \\
      Gemma-4-31B-It & \texttt{google/gemma-4-31b-it} & 1 fps & False & 0 & 1 & vLLM \\
      Qwen3.5-397B-A17B & \texttt{Qwen/Qwen3.5-397B-A17B} & 2 fps & False & 0 & 1 & vLLM \\
      Qwen3.5-122B-A10B & \texttt{Qwen/Qwen3.5-122B-A10B} & 2 fps & False & 0 & 1 & vLLM \\
      Qwen3.5-27B & \texttt{Qwen/Qwen3.5-27B} & 2 fps & False & 0 & 1 & vLLM \\
      Qwen3.5-9B & \texttt{Qwen/Qwen3.5-9B} & 2 fps & False & 0 & 1 & vLLM \\
      \bottomrule
    \end{tabular}
  }
\caption{\revise{\textbf{Inference configurations of the MLLMs evaluated in the main experiment}, including model versions, input frame configurations, decoding hyperparameters, and inference backends. The symbol ``$-$'' indicates that the parameter is not exposed or cannot be configured.}}
  \label{tab:model_config}
\end{table*}

\subsection{Model Licenses}
Table~\ref{tab:model_licenses} summarizes the licenses or access conditions of the models evaluated in this work. Proprietary models and Kimi-k2.5 are accessed through official APIs, while locally deployed open-source models are evaluated according to their released model and code licenses.

\subsection{Computational Resources}
All open-source models were deployed locally in BF16 precision on a computing node equipped with 8$\times$NVIDIA H100 (80GB) GPUs. The evaluation of these local models required approximately 240 GPU hours. All inferences were conducted strictly for academic research and evaluation purposes.

\begin{table}[htbp]
  \centering
  \small
  \renewcommand{\arraystretch}{1.2} 
  
  \resizebox{\columnwidth}{!}{ 
    \begin{tabular}{l c c}
      \toprule
      \textbf{Model} & \textbf{Model License / Access} & \textbf{Code License / Access} \\
      \midrule
      \rowcolor{gray!15} 
      \multicolumn{3}{c}{\textit{Official API Models}} \\
      \midrule
      GPT-5.4 & Official API terms & Official API terms \\
      Gemini-3.1-Pro & Official API terms & Official API terms \\
      Claude-4.6-Sonnet & Official API terms & Official API terms \\
      Kimi-k2.5 & Official API terms & Official API terms \\
      \midrule
      \rowcolor{gray!15} 
      \multicolumn{3}{c}{\textit{Locally Deployed Open-Source Models}} \\
      \midrule
      Gemma-4-31B-It & Apache 2.0 & Apache 2.0 \\
      Qwen3.5-397B-A17B & Apache 2.0 & Apache 2.0 \\
      Qwen3.5-122B-A10B & Apache 2.0 & Apache 2.0 \\
      Qwen3.5-27B & Apache 2.0 & Apache 2.0 \\
      Qwen3.5-9B & Apache 2.0 & Apache 2.0 \\
      \bottomrule
    \end{tabular}
  }
  \caption{Summary of licenses and access conditions for the models evaluated in DVBench.}
  \label{tab:model_licenses}
\end{table}

\subsection{Evaluation Metrics}
\label{sec:appendix_evaluation_metrics}
\paragraph{Exact Match (EM) Tasks.} 
For tasks requiring deterministic and precise answers (e.g., specific numerical values, dates, or concise data facts), we utilize the EM metric. To ensure a robust and fair evaluation against minor formatting variations, both the model's generated outputs and the ground-truth answers undergo standard text normalization (e.g., lowercasing, stripping punctuation, and removing articles) prior to the string matching process.

\paragraph{Choice-Based Tasks (Single and Multiple Choice).} 
For single-choice and multiple-choice questions, we adopt standard accuracy as an evaluation metric. To rigorously evaluate multiple-choice questions, a model's prediction is considered correct if and only if it exactly matches the complete set of ground-truth options (i.e., exact subset match, with no missing or incorrect options selected).

\paragraph{Open-Ended Tasks (Alignment Dimension).} 
For open-ended questions, where models are required to fill in the missing subtitles, we evaluate model outputs from three perspectives: text similarity, visual-text grounding, and human judgment.

\textbf{Text-based Metrics.}
We evaluate the similarity between generated and reference text using \textbf{BLEU-2}~\cite{papineni2002bleu}, \textbf{METEOR}~\cite{banerjee2005meteor}, and \textbf{BERTScore F1}~\cite{zhang2019bertscore}. To ensure reproducibility of the lexical metrics, we compute the scores using standardized Python packages. Specifically, we utilize the NLTK library (\texttt{nltk$\geq$3.6.0}) for tokenization (\texttt{punkt}) and metric computation. \textbf{BLEU-2} evaluates bi-gram overlap between the generated text and the reference answer. To reduce the penalty for short responses with zero n-gram counts, we configure BLEU using NLTK's smoothing function \texttt{method1} (\texttt{SmoothingFunction().method1}). \textbf{METEOR} further incorporates stemming and synonym matching through NLTK's WordNet resources (\texttt{wordnet} and \texttt{omw-1.4}). Beyond lexical matching, \textbf{BERTScore F1} measures semantic similarity between the generated and reference text based on contextual token embeddings. We implement BERTScore using the official \texttt{bert\_score} package (v0.3.13) with \texttt{roberta-large} as the contextual embedding backbone.

\revise{\textbf{Visual Grounding Metrics.}
To further evaluate whether the generated text is grounded in the visual content of the video, we employ \textbf{EMScore} and $\mathbf{EMScore}_{\mathrm{ref}}$. EMScore measures the correspondence between the generated text and the video by jointly considering global video-text similarity and fine-grained frame-word alignment. $\mathrm{EMScore}_{\mathrm{ref}}$ further incorporates the reference text into the evaluation, assessing the generated response with respect to both the visual evidence and the reference semantics.}

\revise{\textbf{Human Evaluation.}
We additionally conduct human evaluation to directly assess the quality of the generated subtitles. Human evaluators rate each response on a five-point scale according to its correctness, consistency with the visual evidence, and faithfulness to the intended message. A score of 1 indicates a response that is incorrect or unsupported by the video, whereas a score of 5 indicates a fully correct and visually grounded response that faithfully conveys the intended meaning. We report the average human rating for each model.}

\subsection{Evaluation Prompt}
\label{sec:appendix_eval_prompt}
\begin{tcolorbox}[
    title={Prompt Template: Exact Match (EM) Task},
    fonttitle=\small\bfseries,
    colbacktitle=gray!70!white,
    coltitle=white,
    colback=gray!5!white,
    colframe=gray!70!white,
    arc=2mm,
    boxrule=0.5pt,
    breakable,
    left=6pt, right=6pt, top=6pt, bottom=6pt
]
\small
\begin{flushleft}
\ttfamily
Based on the video, please answer the following question:\par
\textcolor{blue}{\textit{\{Question Text\}}}\par
Please provide the answer directly in \textless answer\textgreater\textless /answer\textgreater{} without any units, symbols, or explanation. For example, if the answer is 15\% or \$1, output only \textless answer\textgreater 15\textless /answer\textgreater{} or \textless answer\textgreater 1\textless /answer\textgreater.\par
Answer:
\end{flushleft}
\end{tcolorbox}
\vspace{1em}

\begin{tcolorbox}[
    title={Prompt for Multiple-Choice Task (Single Answer)},
    fonttitle=\small\bfseries,
    colbacktitle=gray!70!white,
    coltitle=white,
    colback=gray!5!white,
    colframe=gray!70!white,
    arc=2mm,
    boxrule=0.5pt,
    breakable,
    left=6pt, right=6pt, top=6pt, bottom=6pt
]
\small
\begin{flushleft}
\ttfamily
Based on the video, please answer the following question:\par
\textcolor{blue}{\textit{\{Question Text\}}}\par
A. \textcolor{blue}{\textit{\{Option 1\}}}\par
B. \textcolor{blue}{\textit{\{Option 2\}}}\par
C. \textcolor{blue}{\textit{\{Option 3\}}}\par
D. \textcolor{blue}{\textit{\{Option 4\}}}\par
Output only the answer label (e.g., A, B, C, or D) between \textless answer\textgreater{} and \textless /answer\textgreater{} tags (e.g., \textless answer\textgreater A\textless /answer\textgreater).\par
Answer:
\end{flushleft}
\end{tcolorbox}

\begin{tcolorbox}[
    title={Prompt for Multiple-Choice Task (Multiple Answers)},
    fonttitle=\small\bfseries,
    colbacktitle=gray!70!white,
    coltitle=white,
    colback=gray!5!white,
    colframe=gray!70!white,
    arc=2mm,
    boxrule=0.5pt,
    breakable,
    left=6pt, right=6pt, top=6pt, bottom=6pt
]
\small
\begin{flushleft}
\ttfamily
Based on the video, please answer the following question:\par
\textcolor{blue}{\textit{\{Question Text\}}}\par
A. \textcolor{blue}{\textit{\{Option 1\}}}\par
B. \textcolor{blue}{\textit{\{Option 2\}}}\par
C. \textcolor{blue}{\textit{\{Option 3\}}}\par
D. \textcolor{blue}{\textit{\{Option 4\}}}\par
This question has more than one correct answer. Output answer labels of all correct options answer label between \textless answer\textgreater{} and \textless /answer\textgreater{} tags (e.g., \textless answer\textgreater A;B;C\textless /answer\textgreater).\par
Answer:
\end{flushleft}
\end{tcolorbox}

\begin{tcolorbox}[
    title={Prompt for Alignment Dimension},
    fonttitle=\small\bfseries,
    colbacktitle=gray!70!white,
    coltitle=white,
    colback=gray!5!white,
    colframe=gray!70!white,
    arc=2mm,
    boxrule=0.5pt,
    breakable,
    left=6pt, right=6pt, top=6pt, bottom=6pt
]
\small
\begin{flushleft}
\ttfamily
Please analyze the video and the surrounding subtitle context to identify the missing segment:\par
Subtitle Context: \\
\textcolor{blue}{\textit{\{Subtitle\}}} \par

Based on the video content and the context of the subtitles, please fill in the narration for \textit{[Insert Subtitle Here]}, and output the filled content directly between the \textless answer\textgreater{} and \textless /answer\textgreater{} tags.\par
Answer:
\end{flushleft}
\end{tcolorbox}

\section{Timeline Question Analysis}
\label{sec:appendix_timeline_sampling}

To further investigate whether the difficulty of \textit{Timeline} questions
is primarily caused by sparse frame sampling, we evaluate these questions
under different frame conditions available to each model. The results (Table~\ref{tab:timeline_sampling}) indicate that frame sampling can affect individual
\textit{Timeline} performance, but increasing the frame budget or sampling
rate does not consistently resolve the difficulty. We therefore consider
sparse sampling to be one possible contributor to the low performance on
\textit{Timeline} questions, rather than the sole cause.

\begin{table}[!t]
\centering
\scriptsize
\setlength{\tabcolsep}{3.5pt}
\renewcommand{\arraystretch}{1.1}

\resizebox{\columnwidth}{!}{
\begin{tabular}{lcccc}
\toprule
\textbf{Model}
& \textbf{64 frames}
& \textbf{0.5 fps}
& \textbf{1 fps}
& \textbf{Original setting} \\
\midrule

Gemini-3.1-Pro
& 37.5 & - & 37.5 & 37.5 (1 fps) \\

Claude-4.6-Sonnet
& 50.0 & - & - & 37.5 (100 frames) \\

GPT-5.4
& 37.5 & - & - & 37.5 (64 frames) \\

Qwen3.5-397B-A17B
& 50.0 & 50.0 & 50.0 & 62.5 (2 fps) \\

Qwen3.5-122B-A10B
& 25.0 & 25.0 & 25.0 & 25.0 (2 fps) \\

Qwen3.5-27B
& 37.5 & 37.5 & 37.5 & 50.0 (2 fps) \\

Qwen3.5-9B
& 37.5 & 37.5 & 50.0 & 12.5 (2 fps) \\

\bottomrule
\end{tabular}
}

\caption{
\textbf{Timeline performance under different frame sampling conditions.}
We report accuracy under different sampling rates
where supported, and each model's original inference setting. The sampling rate or frame budget used in each model's
original setting is shown in parentheses.
}
\label{tab:timeline_sampling}
\end{table}


\section{Qualitative Case Studies}
\label{sec:appendix_cases}

To provide a more intuitive understanding of the models' capabilities and failure modes, we present several qualitative case studies across different evaluation dimensions.

\subsection{Animation Evaluation Cases}
\label{sec:appendix_case_animation}

\begin{figure*}[!h]
    \centering
    \includegraphics[width=\linewidth]{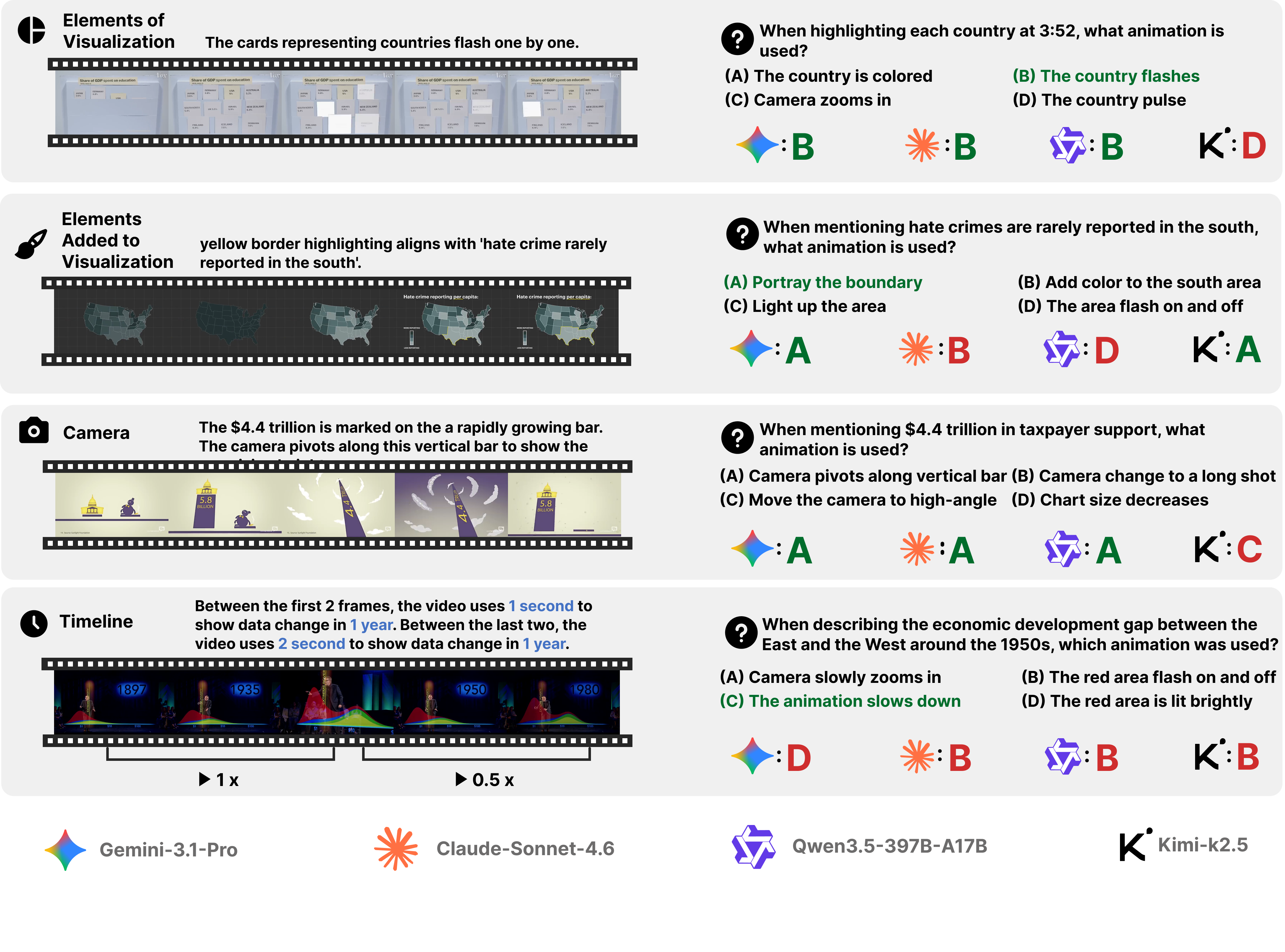}
    \caption{\textbf{Cases of Animation QA pairs.} We present questions of four editorial layers here. As we can see, questions of animation on Timeline layer (e.g., slow down, speed up, and pause in visualization animation) are most challenging for models.}
    \label{fig:animation_case}
\end{figure*}

We present qualitative examples across the four editorial layers of animation in Fig.~\ref{fig:animation_case}. The cases highlight a critical vulnerability: while MLLMs can generally identify explicit visual transformations (e.g., color changes or added elements), they consistently struggle with the Timeline layer. Tasks requiring the perception of non-linear animation pacing, such as identifying when a chart speeds up, slows down, or pauses, prove exceptionally challenging.

\subsection{Chart Reasoning Evaluation Cases}
\label{sec:appendix_case_reasoning}
\begin{figure*}[!h]
    \centering
    \includegraphics[width=\linewidth]{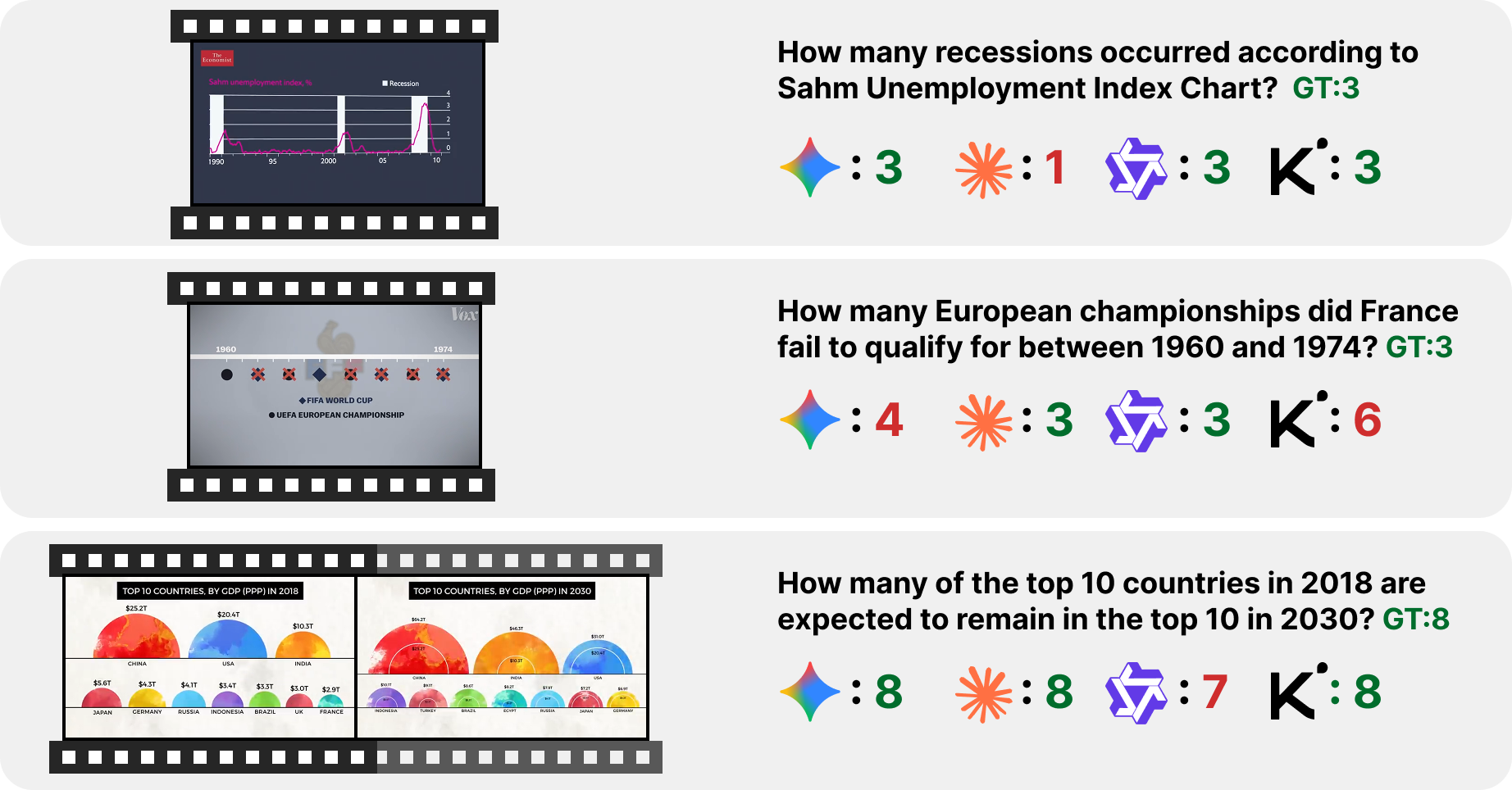}
    \caption{\textbf{Cases of Chart Reasoning QA pairs.} We illustrate three Aggregation questions. While the first two questions can be answered using a single keyframe, the third requires integrating information across two discrete frames. Successfully answering these queries requires models to accurately ground target chart elements across single or multiple frames before performing specific numerical calculations.}
    \label{fig:reasoning_case}
\end{figure*}

Fig.~\ref{fig:reasoning_case} illustrates qualitative examples from the Aggregation questions in Chart Reasoning. 

\subsection{Alignment Analysis}
\label{sec:appendix_case_alignment}
\begin{figure*}[!h]
    \centering
    \includegraphics[width=\linewidth]{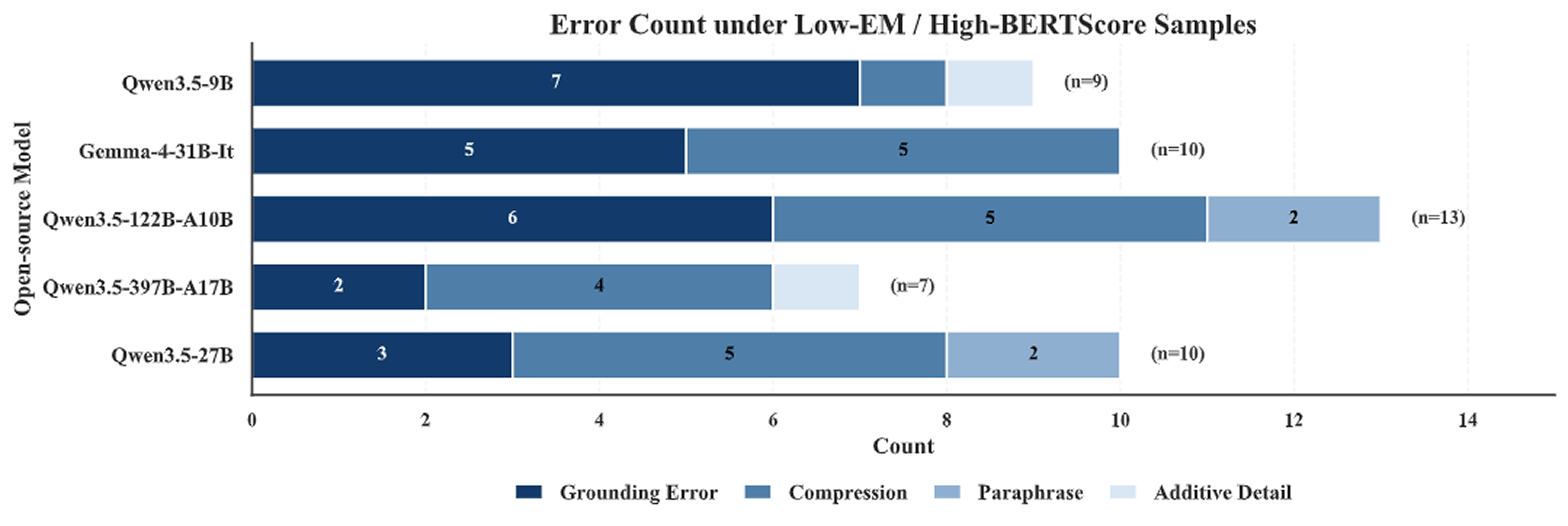}
    \caption{\textbf{Counts of error types for the low-EM / high-BERTScore answers across open-source models.} Each horizontal stacked bar shows the number of examples in the slice (defined per model as BLEU-2 ≤ model median, METEOR ≤ model median, and BERTScore ≥ model median), grouped by four error types. Per-row totals are shown to the right as (n=).}
    \label{fig:alignment_error_count}
\end{figure*}
To investigate the specific reasons why open-source models significantly underperform proprietary models on EM metrics (BLEU-2 and METEOR), we conduct a qualitative analysis of the \emph{low-EM / high-BERTScore} cases generated by open-source models. We define such a case at the model level as a sample whose EM scores are both below the model's median, while its BERTScore is at or above the median. We identify four error types: \textbf{(1) Grounding Error}: the predicted answer is factually inconsistent with the gold answer or the visual evidence. \textbf{(2) Compression}: the prediction preserves the core meaning but drops important details. \textbf{(3) Paraphrase}: the prediction is semantically equivalent to the gold answer, with only surface-form variation. \textbf{(4) Additive Detail}: the prediction preserves the core meaning but adds extra details, numbers, or assertions that are not present in the gold answer.

As shown in Fig.~\ref{fig:alignment_error_count}, most examples fall into Grounding Error and Compression, indicating outputs that are semantically similar to references but omit or compress key information, causing a low exact-match score. Model-specific differences appear: e.g., Qwen3.5-9B has a larger share of Grounding Error, while other models show relatively higher compression. Table~\ref{tab:appendix_alignment_case} presents representative cases, illustrating typical error patterns.

\begin{table*}[!h]
\centering
\footnotesize
\renewcommand{\arraystretch}{1.15}

\begin{tabularx}{\textwidth}{
    l
    >{\raggedright\arraybackslash}X
    c
}
\toprule
\textbf{Model} & \textbf{Model Response} & \textbf{Error Type} \\
\midrule

\rowcolor{gray!15}
\multicolumn{3}{c}{
\parbox{0.96\textwidth}{
\centering
\textbf{Case 1:} there changed the whole pattern of the world it became a divided world
the world view you grew up with with a rich West here and with poor Africa and Asia here
}}
\\
\midrule

Gemma-4-31B-It
& and then in 1948 the population grew and the incomes increased even more and the world looked like this
& \cellcolor{red!15}Grounding error \\
\hline

Qwen3.5-397B-A17B
& the shape of the world changed
& \cellcolor{gray!15}Compression \\
\hline

Qwen3.5-122B-A10B
& in the 1970s the population grew here and the incomes increased here
& \cellcolor{red!15}Grounding error \\
\hline

Qwen3.5-27B
& but in the 1980s the growth of China they start to get out of poverty they close this Gap and they are coming over on this side and the shape of the world change again and look now Africa is following here we are today in a completely new shape of the world for those of you who still think about the world as the rich and the poor
& \cellcolor{purple!15}Additive detail \\
\hline

Qwen3.5-9B
& and the shape of the world change again
& \cellcolor{gray!15}Compression \\

\midrule

\rowcolor{gray!15}
\multicolumn{3}{c}{
\parbox{0.96\textwidth}{
\centering
\textbf{Case 2:} But the IRS found that from 2008 to 2010 Americans paid only 81.7 percent
of the taxes they really owed. Assuming that rate is the same today,
That's about \$660 billion dollars in tax dodging each year.
}}
\\
\midrule

Gemma-4-31B-It
& But that's not all that's owed. According to an IRS study, about 18 percent of taxes owed are never paid. That's 660 billion dollars.
& \cellcolor{blue!15}Paraphrase \\

\bottomrule
\end{tabularx}

\caption{
\textbf{Representative alignment cases illustrating different error types.}
For each case, the gray-highlighted text at the top shows the gold answer,
followed by model-generated responses and their corresponding error types.
}
\label{tab:appendix_alignment_case}
\end{table*}
\section{AI Assistants Usage Declaration}
AI assistants were utilized to facilitate several stages of this work, including dataset construction, coding support, figure design, and language polishing. Crucially, all AI-generated outputs underwent rigorous manual verification and extensive revision by the authors to ensure accuracy and quality.

\end{document}